\documentclass[letterpaper,journal]{IEEEtran}

\usepackage{amsmath,amsfonts}
\usepackage{newtxtext,newtxmath}
\usepackage{algorithmic}
\usepackage{algorithm}
\usepackage{array}
\usepackage{booktabs}
\usepackage{multirow}
\usepackage[caption=false,font=normalsize,labelfont=sf,textfont=sf]{subfig}
\usepackage{textcomp}
\usepackage{stfloats}
\usepackage{url}
\usepackage{verbatim}
\usepackage{graphicx}
\usepackage{cite}

\begin{document}

\title{AIR: Amortized Image Reconstruction Framework for Self-Supervised Feed-Forward 2D Gaussian Splatting}

\author{Zhaojie~Zeng,
        Yuesong~Wang,
        Yawei~Luo,
        Tao~Guan
\thanks{Zhaojie Zeng, Yuesong Wang, and Tao Guan are with the School of Computer Science and Technology, Huazhong University of Science and Technology, Wuhan, China (e-mail: \{zhaojiezeng, yuesongwang, qd\_gt\}@hust.edu.cn).}%
\thanks{Yawei Luo is with the School of Software Technology, Zhejiang University, Hangzhou, Zhejiang (e-mail: yaweiluo@zju.edu.cn).}%
\thanks{Corresponding author: Yuesong Wang (e-mail: yuesongwang@hust.edu.cn).}}

\markboth{Preprint}%
{Zeng \MakeLowercase{\textit{et al.}}: AIR}


\maketitle

\begin{abstract}
2D Gaussian splatting provides an efficient explicit representation for image reconstruction, but existing methods still require costly per-image iterative optimization or rely on handcrafted priors for primitive allocation.
We present AIR, a self-supervised feed-forward framework that amortizes iterative Gaussian fitting into a single network pass, eliminating per-image test-time optimization.
AIR adopts a stage-wise residual architecture that progressively predicts additional Gaussian primitives from reconstruction residuals, together with an explicit Stage Control mechanism that activates new primitives only in under-reconstructed regions.
A Predict--Optimize--Distill training strategy stabilizes multi-stage prediction by distilling short-horizon optimized Gaussian increments back into the predictor.
The stabilized predictor is then jointly finetuned across stages and equipped with an image-adaptive quantizer for compact Gaussian storage.
Experiments on Kodak and DIV2K show that AIR achieves better reconstruction quality than representative Gaussian-based baselines while reducing encoding time to 160--300\,ms.
Code: \url{https://github.com/whoiszzj/AIR.git}
\end{abstract}


\section{Introduction}

\IEEEPARstart{I}{mage} reconstruction studies how to encode a target image into an alternative representation and recover its visual content through an efficient decoding or rendering process~\cite{ongie2020deep,xie2022neural}. 
Instead of treating an image only as a dense grid of pixel values, image reconstruction methods seek compact neural or explicit representations that can faithfully model visual content while supporting efficient storage, rendering, or transmission~\cite{essakine2024we}. 
This makes image reconstruction a fundamental problem for compact visual representation, efficient rendering, and image-based communication.

One representative direction is to model images with neural networks, especially coordinate-based implicit neural representations~\cite{klocek2019hypernetwork, sitzmann2020siren}.
These methods fit a neural function that maps spatial coordinates to color values and can achieve high-quality reconstruction.
However, recovering an image requires dense network evaluation over all pixel locations, which leads to a relatively high decoding cost.
Recently, explicit 2D Gaussian representations~\cite{gaussianimage,ImageGS, zhu2025large} have emerged as an efficient alternative.
By decomposing an image into a set of Gaussian primitives and rendering them through differentiable splatting, Gaussian-based methods provide fast decoding while maintaining strong reconstruction fidelity.

Despite their efficient decoding, existing 2D Gaussian image reconstruction methods are still limited by their reliance on image-specific primitive construction or optimization.
GaussianImage~\cite{gaussianimage} optimizes Gaussian primitives independently for each image, which yields high-quality results but incurs costly per-image fitting.
Image-GS~\cite{ImageGS} improves the optimization process, yet still follows the same image-specific optimization paradigm.
Instant-GI~\cite{Instant-GI} reduces the encoding cost by learning to predict an initial Gaussian set, but this prediction mainly serves as an initialization.
High-fidelity reconstruction still typically requires subsequent per-image refinement, leaving the image-specific optimization process only semi-amortized~\cite{AmortizedOptimization}.
In addition, its adaptive prediction of the number of Gaussians and primitive attributes relies on substantial handcrafted priors.
Although these priors can help organize the Gaussian layout, they inevitably introduce inductive bias and prevent the overall pipeline from being fully end-to-end.

\begin{figure}[t!]
    \centering
    \includegraphics[width=0.95\linewidth]{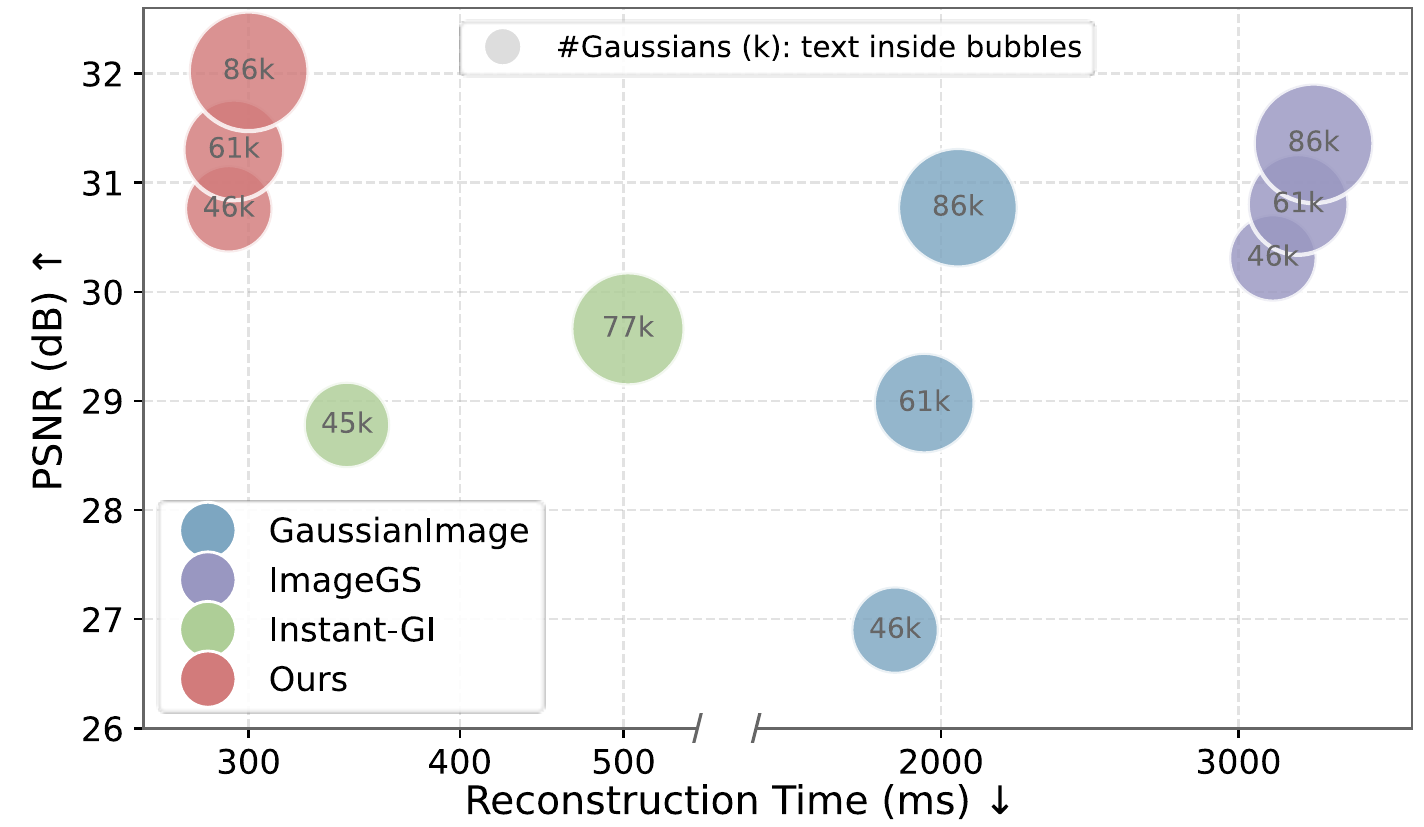}
    \caption{
        Comparison of reconstruction time and PSNR on the DIV2K validation set.
        The x-axis uses a broken logarithmic scale for visualization.
        AIR achieves better reconstruction quality with substantially lower reconstruction time than GaussianImage, Image-GS, and Instant-GI.
        }
    \label{fig:teaser}
\end{figure}

In this paper, we propose \textbf{AIR}, an \emph{Amortized Image Reconstruction} framework for self-supervised feed-forward 2D Gaussian splatting. 
Our goal is to amortize the image-specific Gaussian fitting process itself, rather than merely learning a better initialization. 
AIR reconstructs an image through stage-wise residual Gaussian prediction. 
Starting from an empty Gaussian set, each stage predicts additional Gaussian primitives from the current reconstruction residual and accumulates them into the existing set. 
The final representation is therefore formed progressively, with later stages correcting the residual errors left by earlier stages.

Training this feed-forward predictor is non-trivial, since the mapping from image space to Gaussian space is highly under-constrained, and direct supervision with rendering loss may lead to unstable optimization.
We therefore introduce a Predict--Optimize--Distill (POD) training strategy.
At each stage, the predictor outputs a stage-specific Gaussian primitive set that is added to the accumulated representation.
We create a detached copy of this set and refine only the copy through a few differentiable rendering optimization steps.
The refined set is then used as a Gaussian-space regression target to supervise the original prediction and update the predictor parameters.

After the POD training strategy stabilizes the multi-stage predictor, AIR is then finetuned with direct rendering supervision on the whole Gaussian set and further equipped with an image-adaptive quantizer for compact Gaussian storage without per-image post-optimization.

Experiments on Kodak and DIV2K show that AIR achieves higher reconstruction quality than prior 2D Gaussian methods while reducing encoding time to 160--300\,ms, without relying on per-image optimization or handcrafted priors.

The main contributions are as follows:
\begin{itemize}
    \item We propose AIR, a self-supervised feed-forward framework for amortized 2D Gaussian image reconstruction, which generates Gaussian representations directly at inference time without image-specific iterative optimization.
    
    \item We formulate image reconstruction as stage-wise residual Gaussian prediction with explicit stage control, progressively adding Gaussian primitives only to regions that have not reached the prescribed fidelity target.
    
    \item We develop a training strategy tailored to feed-forward Gaussian prediction, where POD pretraining stabilizes Gaussian-space supervision and subsequent rendering finetuning improves the final accumulated reconstruction, while an image-adaptive quantizer enables compact storage without per-image calibration.
\end{itemize}
\section{Related Work}

\subsection{Implicit Neural Image Representation}

Coordinate-based implicit neural representations (INRs) model visual signals as continuous neural functions, mapping 2D image coordinates to RGB values for image representation~\cite{klocek2019hypernetwork,chen2021learning,dupont2021coin,dupont2022coinpp} or 3D positions and view directions to radiance fields for novel-view synthesis~\cite{NeRF, barron2022mipnerf360}. 
To better represent high-frequency details, existing methods introduce positional encoding, Fourier features, periodic activations, or spectral analysis of coordinate networks~\cite{tancik2020fourier,sitzmann2020siren,benbarka2022seeing}. 
Other works improve fitting efficiency and signal fidelity through adaptive coordinate decomposition, multi-resolution feature grids, hash encodings, band-limited coordinate networks, or wavelet-based activations~\cite{martel2021acorn,muller2022instantngp,lindell2022bacon,saragadam2023wire}. 
INRs have also been explored for compact signal representation and compression, often with meta-learning or sparsification to reduce per-sample fitting cost~\cite{strumpler2022implicit,lee2021meta,schwarz2022meta}. 

Despite their flexibility, INR-based image representations usually require image-specific optimization, and decoding an image involves dense network evaluation over pixel coordinates. 
AIR instead adopts an explicit 2D Gaussian representation, which reconstructs images through efficient differentiable splatting and further amortizes Gaussian fitting into feed-forward residual prediction.

\subsection{Gaussian-based Image Representation}

Gaussian splatting provides an explicit and efficient representation for visual reconstruction. 
Following the success of 3D Gaussian Splatting~\cite{kerbl20233dgaussians}, GaussianImage~\cite{gaussianimage} introduces 2D Gaussian splatting for image representation and compression, establishing a primitive-based rendering paradigm that achieves fast decoding with strong reconstruction quality. 
Subsequent works improve this paradigm by introducing better initialization and error-driven allocation strategies~\cite{ImageGS,liang2025structure,li2025gaussianimagepp}. 
By using image structure or rendering distortion as guidance, these methods allocate Gaussians more effectively across different image regions. 
However, the allocation still follows a pre-defined number of Gaussians or a densification schedule, and the final representation is obtained through long per-image iterative optimization. 

Instant-GI~\cite{Instant-GI} takes a further step toward adaptive image representation by predicting an initial Gaussian set with a network and dynamically adjusting the number of Gaussians according to image complexity. 
Nevertheless, its adaptive primitive allocation relies on PPM pseudo labels generated by a handcrafted initialization pipeline, and high-fidelity reconstruction still typically requires subsequent per-image refinement. 
In contrast, AIR aims to amortize the Gaussian fitting process itself and predicts Gaussian representations in a self-supervised feed-forward manner.

\subsection{Amortized Optimization and Iterative Refinement}

AIR is related to amortized optimization and learned iterative refinement.
Amortized optimization replaces repeated test-time optimization with a learned predictor~\cite{AmortizedOptimization}, while learned iterative refinement progressively corrects an initial prediction using feedback signals~\cite{marino2018iterative,adler2018learned,putzky2017recurrent}.
In Gaussian splatting, \cite{xu2025resplat,chen2026gifsplat} follows a related idea by recurrently updating an initial 3D Gaussian representation using rendering errors as feedback.
In contrast, AIR focuses on single-image 2D Gaussian reconstruction and predicts residual Gaussian increments stage by stage, adding new primitives to under-reconstructed regions rather than repeatedly updating a fixed Gaussian set.

AIR is also inspired by ~\cite{wu2025predict}, which refines a prediction through short optimization and distills the optimized result back to the predictor.
AIR transfers this idea to residual 2D Gaussian reconstruction: each stage predicts a new subset of Gaussian primitives, optimizes a detached copy of this subset with the rendering loss, and uses the optimized subset as a Gaussian-space regression target.
\section{Method}
\label{sec:method}

\subsection{Preliminaries}
\label{sec:preliminaries}

Following~\cite{gaussianimage}, we consider image reconstruction as an optimization over a set of 2D Gaussians. Let $\mathcal{R}(\cdot)$ denote the differentiable renderer function. Given a Gaussian set $G$, the rendered image is defined as $I_{render} = \mathcal{R}(G)$. For a target image $I_{gt}$, the goal is to estimate the optimal Gaussian set $G^{\star}$ by minimizing the photometric reconstruction loss:

\begin{equation}
    G^{\star} \in \arg\min_{G} \mathcal{L}_{render}\!\left(I_{render}, I_{gt}\right),
\end{equation}

To further account for representation compactness, we seek the smallest Gaussian set that satisfies a prescribed fidelity target:
\begin{equation}
    \label{eq:constrained_optimization_problem}
    G^{\star} \in 
    \arg\min_{G} |G|
    \quad
    \mathrm{s.t.}
    \quad
    \mathcal{L}_{render}\!\left(\mathcal{R}(G), I_{gt}\right) \leq \epsilon .
\end{equation}

GaussianImage~\cite{gaussianimage} reduces the 3D Gaussian primitive of 3DGS to a purely 2D representation. The $n$-th primitive is parameterized by a 2D center $\boldsymbol{\mu}_n \in \mathbb{R}^2$, a covariance matrix $\boldsymbol{\Sigma}_n \in \mathbb{R}^{2 \times 2}$, and an RGB color $\boldsymbol{c}_n \in \mathbb{R}^3$ that absorbs opacity. To ensure positive definiteness, $\boldsymbol{\Sigma}_n$ is expressed in rotation-scale form:
\begin{equation} 
    \boldsymbol{\Sigma}_n = (\boldsymbol{R}_n \boldsymbol{S}_n)(\boldsymbol{R}_n \boldsymbol{S}_n)^T,
\end{equation} 
where $\boldsymbol{R}_n$ is a 2D rotation matrix determined by an angle $\theta_n$, and $\boldsymbol{S}_n = \mathrm{diag}(s_{n_1}, s_{n_2})$ controls the Gaussian size.

Since all Gaussians are defined on a common image plane, no depth ordering is required, and the color at pixel $p$ can be computed as a weighted aggregation:
\begin{equation}
    \label{eq:rendering_equation}
    \boldsymbol{C}_p = \sum_{n \in \mathcal{N}(p)} \boldsymbol{c}_n \alpha_{n,p},
    \qquad
    \alpha_{n,p} = \exp\!\left(-\tfrac{1}{2}\, \boldsymbol{d}_{n,p}^T \boldsymbol{\Sigma}_n^{-1} \boldsymbol{d}_{n,p}\right),
\end{equation}
where $\mathcal{N}(p)$ is the set of Gaussians covering pixel $p$ and $\boldsymbol{d}_{n,p}$ is the offset from $\boldsymbol{\mu}_n$ to pixel $p$.

To accelerate GaussianImage~\cite{gaussianimage}, Instant-GI~\cite{Instant-GI} adopts a semi-amortized~\cite{AmortizedOptimization} formulation with learned initialization: a network predicts an initial Gaussian set, after which only a few optimization steps are required to reach a comparable reconstruction, substantially shortening the encoding time.

Despite this acceleration, Instant-GI still has two important limitations. First, to determine where and how many Gaussians to allocate, Instant-GI introduces a Position Probability Map (PPM) and combines it with Floyd--Steinberg dithering~\cite{franchini2019stochastic} to sample Gaussian centers according to local image complexity. However, this PPM-based sampling process is non-differentiable, so the network must be trained with pseudo-label supervision rather than directly from the rendering loss, which hinders fully self-supervised training on large image corpora. Second, the pipeline relies on several handcrafted priors, including quadtree-based density estimation from image gradients and Gaussian initialization through Delaunay triangulation~\cite{watson1981delaunay} followed by ellipse fitting. Despite these priors, the predicted initialization still falls short of the target image in both SSIM and perceptual quality, and iterative refinement remains necessary to achieve competitive reconstructions.
\subsection{Overview}
\label{sec:overview}

\begin{figure*}[!t]
    \centering
    \includegraphics[width=\linewidth]{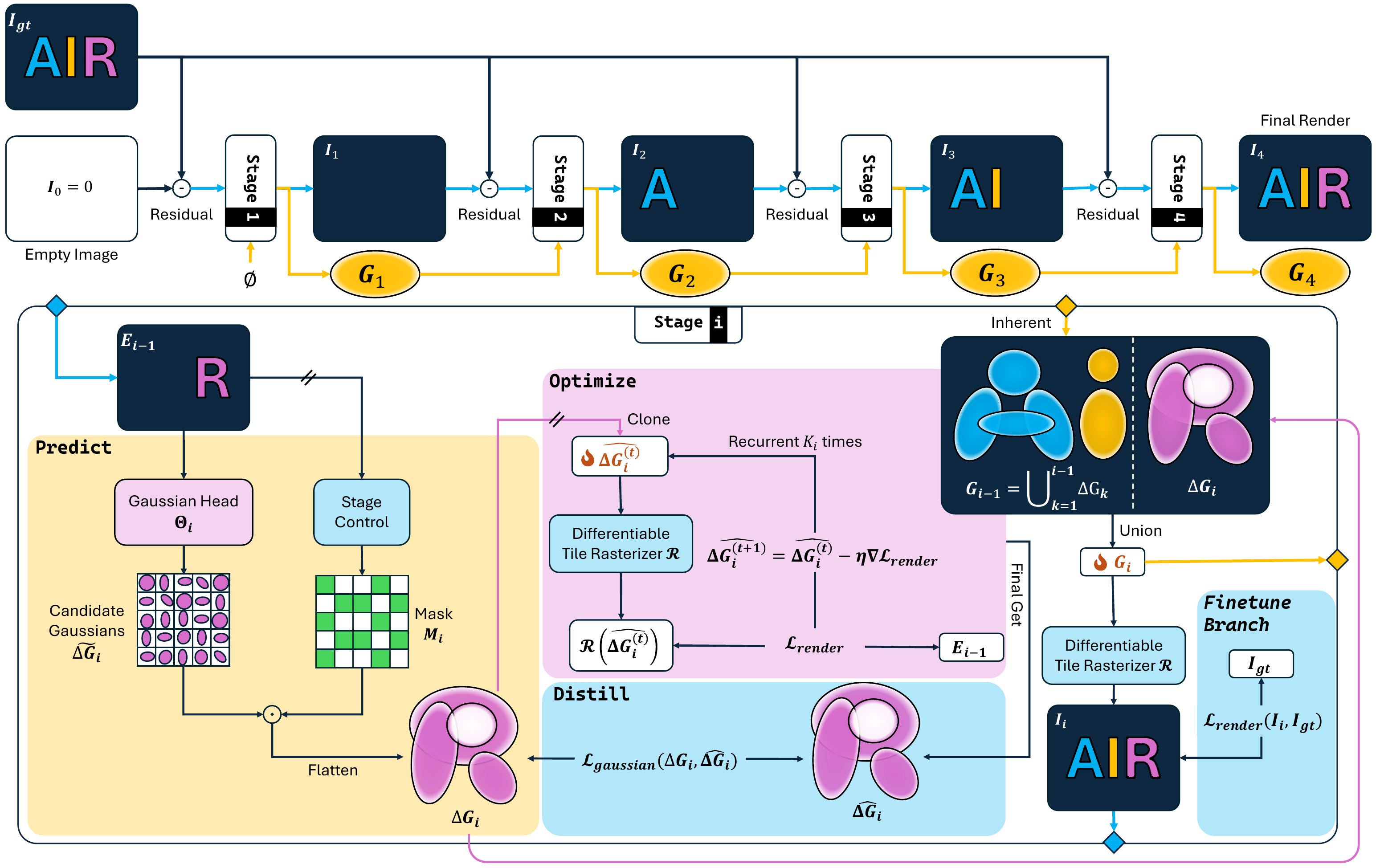}
    \caption{
        Overview of AIR. AIR amortizes the iterative optimization of GaussianImage through self-supervised stage-wise residual prediction.
        At each stage, the network predicts Gaussian increments from the current reconstruction residual, while an explicit stage-control mechanism activates additional primitives only in under-reconstructed regions.
        During POD pretraining, a short-horizon optimizer refines a detached copy of each predicted increments and distills the refined increments back as Gaussian-space supervision.
        After POD stabilizes the predictor, cross-stage rendering finetuning supervises the accumulated reconstructions directly, promoting cross-stage coordination for the final output.
        }
    \label{fig:pipeline}
\end{figure*}

To address the above limitations of Instant-GI, we propose \textbf{AIR}, a fully self-supervised feed-forward framework that amortizes the iterative optimization of GaussianImage.
First, we design a stage-wise residual Gaussian predictor (Sec.~\ref{sec:stage_wise_residual_gaussian_prediction}) that progressively predicts Gaussians across stages and uses an explicit stage-control mechanism to determine where additional Gaussians are still needed.
This design removes the dependence on both pseudo-label supervision and handcrafted priors in Instant-GI, replacing them with residual signals derived from the current reconstruction, thereby making large-scale self-supervised training practical.
Second, considering the Gaussian parameter space is highly under-constrained and training each stage with image-space rendering loss alone is prone to divergence, we introduce a Predict--Optimize--Distill (POD) training strategy (Sec.~\ref{sec:pod_training}).
The newly added Gaussians are first refined by a short-horizon optimizer, and the refined result is then distilled back into the predictor as a better-conditioned regression target, stabilizing end-to-end pretraining.
Finally, after the predictor has been stabilized, we further develop a finetuning-and-quantization scheme (Sec.~\ref{sec:finetune_and_quantization}) to refine the reconstruction and directly quantize the predicted Gaussian representation.
During finetuning, the costly short-horizon optimization used in POD is removed, and the model is trained with direct image-space supervision, reducing the per-iteration training cost.
Meanwhile, learned quantization compresses the Gaussian representation for storage and transmission.
The pipeline is illustrated in Fig.~\ref{fig:pipeline}.

\subsection{Stage-wise Residual Gaussian Prediction}
\label{sec:stage_wise_residual_gaussian_prediction}

One of the key contributions of Instant-GI is its ability to allocate Gaussians adaptively according to local image complexity, which relies on a Position Probability Map supervised by a pseudo-label.
However, generating this pseudo-label already bakes in strong priors, such as gradient-driven quadtree partitioning, and obtaining it requires running GaussianImage for many iterations.
Scaling such a pipeline to larger training corpora is therefore expensive and inevitably injects additional inductive bias.
To address these drawbacks, we propose a hierarchical network built on residual correction (Sec.~\ref{sec:residual_design}), where each stage predicts new Gaussian primitives from the reconstruction produced by the previous stage.
Together with a Stage Control module (Sec.~\ref{sec:stage_control}), the network adaptively allocates Gaussian primitives under an explicit quality constraint.
This follows the constrained reconstruction objective in Eq.~\ref{eq:constrained_optimization_problem}: using as few Gaussians as possible while satisfying a prescribed fidelity target.

\subsubsection{Residual Design}
\label{sec:residual_design}

In the rendering formulation of Eq.~\ref{eq:rendering_equation}, color parameters are allowed to take signed values and no clamping is applied.
Each pixel color is therefore computed as an order-independent sum of the contributions from all covering Gaussians.
This makes the renderer $\mathcal{R}(\cdot)$ additive over disjoint Gaussian sets.
For any two disjoint sets $G_a$ and $G_b$, we have
\begin{equation}
    \mathcal{R}(G_a \cup G_b)
    =
    \mathcal{R}(G_a)
    +
    \mathcal{R}(G_b).
    \label{eq:render_additive}
\end{equation}

This additivity allows reconstruction to be carried out in a stage-wise, incremental fashion.
Let $G_i$ be the accumulated Gaussian set after stage $i$, and let $\Delta G_i$ denote the Gaussians newly introduced at this stage.
The accumulated set is updated as
\begin{equation}
    G_i = G_{i-1} \cup \Delta G_i .
\end{equation}
The rendered image at stage $i$ then decomposes as
\begin{equation}
    I_i
    =
    \mathcal{R}(G_i)
    =
    \mathcal{R}(G_{i-1})
    +
    \mathcal{R}(\Delta G_i)
    =
    I_{i-1}
    +
    \mathcal{R}(\Delta G_i).
\end{equation}

We exploit this decomposition to formulate reconstruction as a sequence of residual corrections.
Instead of regressing the full Gaussian set at once, each stage predicts the Gaussian increment required to explain the remaining error.
Given the current reconstruction $I_{i-1}$, we define the residual as
\begin{equation}
    E_{i-1} = I_{gt} - I_{i-1}.
\end{equation}
We initialize the process with $G_0=\varnothing$ and
\begin{equation}
    I_0 = \mathcal{R}(G_0) = \mathbf{0},
\end{equation}
which gives $E_0=I_{gt}$.
Thus, the first stage directly models the target image, while later stages only model the residual content that has not yet been explained.

The network at stage $i$ takes this residual as input and predicts additional Gaussian primitives:
\begin{equation}
    \Delta G_i = f_{\theta_i}(E_{i-1}).
\end{equation}

\begin{figure}[t!]
  \centering
  \includegraphics[width=\linewidth]{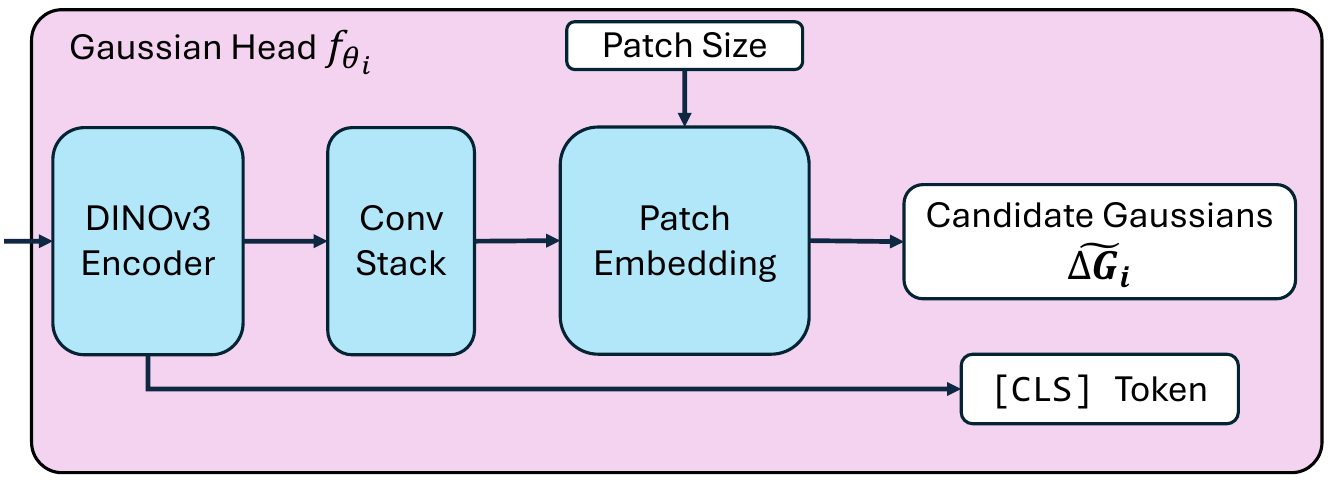}
  \caption{Network architecture for stage-wise residual Gaussian prediction.}
  \label{fig:network}
\end{figure}

The detailed network architecture $f_{\theta_i}$ is illustrated in Fig.~\ref{fig:network}, with implementation details provided in Sec.~\ref{sec:implementation} and the supplementary material.

The newly predicted increments are expected to render the residual component itself, leading to the natural residual rendering objective
\begin{equation}
    \mathcal{L}_{res}^{(i)}
    =
    \mathcal{L}_{render}
    \!\left(
        \mathcal{R}(\Delta G_i),\,
        E_{i-1}
    \right).
    \label{eq:residual_rendering_objective}
\end{equation}

Compared with predicting all Gaussians at once, this formulation decomposes the reconstruction problem along the error signal itself, so that each $f_{\theta_i}$ only needs to model the remaining error rather than the full image content.
By assigning each stage a smaller and more focused residual target, the formulation reduces the learning burden of individual predictors and makes progressive Gaussian reconstruction easier to optimize.

\subsubsection{Stage Control}
\label{sec:stage_control}

The residual formulation above provides a dense spatial signal that indicates where the current reconstruction remains insufficient, making it a natural basis for stage-wise primitive allocation.
Rather than relying on the non-differentiable PPM pseudo-label used by Instant-GI for Gaussian density allocation, we introduce an explicit quality-constrained Stage Control mechanism that activates new Gaussians according to the reconstruction quality of each spatial region.

At each stage, new Gaussians are introduced only in regions whose reconstruction quality has not yet reached a prescribed fidelity target.
This fidelity-guided allocation serves as a greedy approximation to the constrained reconstruction objective in Eq.~\ref{eq:constrained_optimization_problem}: by skipping primitive allocation in regions that already satisfy the quality target, the mechanism avoids inter-stage redundancy and promotes a more compact representation at the global level.
This design is related to content-adaptive resource allocation in image compression and visual recognition, where coding bits~\cite{song2021variable,minnen2017spatially} or computational effort~\cite{figurnov2017spatially} are spatially distributed according to local content characteristics.
Rather than adapting coding or computational resources, AIR adapts the representation itself, activating additional Gaussian primitives only where the current reconstruction still violates the prescribed fidelity target.

Concretely, given the reconstruction $I_{i-1}$ from the previous stage, we compute both a patch-wise PSNR map and a local SSIM map:
\begin{equation}
  \mathbf{Q}^{psnr}_{i-1}
  =
  \mathcal{Q}_{psnr}(I_{gt}, I_{i-1}),
  \qquad
  \mathbf{Q}^{ssim}_{i-1}
  =
  \mathcal{Q}_{ssim}(I_{gt}, I_{i-1}).
\end{equation}
We use PSNR and SSIM as complementary quality measures: PSNR captures patch-wise reconstruction fidelity, while SSIM captures local structural similarity that is more aligned with perceptual image quality~\cite{wang2004image}.

Based on these quality maps, we define the activation mask as
\begin{equation}
  \mathbf{M}_{i}(u)=
  \mathbb{I}\!\left[
    \mathbf{Q}^{psnr}_{i-1}(u) < \tau_{psnr}
    \;\lor\;
    \mathbf{Q}^{ssim}_{i-1}(u) < \tau_{ssim}
  \right],
  \label{eq:stage_mask}
\end{equation}
where $u$ denotes a spatial location.
Thus, $\mathbf{M}_{i}$ selects spatial regions that fail to meet the prescribed fidelity target, restricting new Gaussian allocation to under-reconstructed areas.

To apply this allocation policy, we distinguish the dense Gaussian prediction from the effective activated subset.
Since each spatial token produces one Gaussian, let $\widetilde{\Delta G}_i$ denote the dense candidate Gaussian set predicted by stage $i$:
\begin{equation}
  \widetilde{\Delta G}_i = f_{\theta_i}(E_{i-1}).
\end{equation}
The effective current-stage Gaussian subset is obtained by applying the activation mask:
\begin{equation}
  \Delta G_i = \mathbf{M}_{i} \odot \widetilde{\Delta G}_i,
  \label{eq:gated_gaussian_prediction}
\end{equation}
where $\odot$ denotes patch-wise masking.
As illustrated in Fig.~\ref{fig:pipeline}, this mechanism activates later-stage Gaussians only at under-reconstructed spatial tokens. 
In practice, we set $\tau_{psnr}=35\;\mathrm{dB}$ and $\tau_{ssim}=0.95$ to define a high-fidelity operating regime.
Since new Gaussians are activated only where the quality scores fall below these thresholds, increasing the thresholds imposes a stricter fidelity target and activates more primitives, while decreasing them yields a more compact representation.

\subsection{Predict--Optimize--Distill Training}
\label{sec:pod_training}

\begin{figure}[t!]
    \centering
    \includegraphics[width=0.9\linewidth]{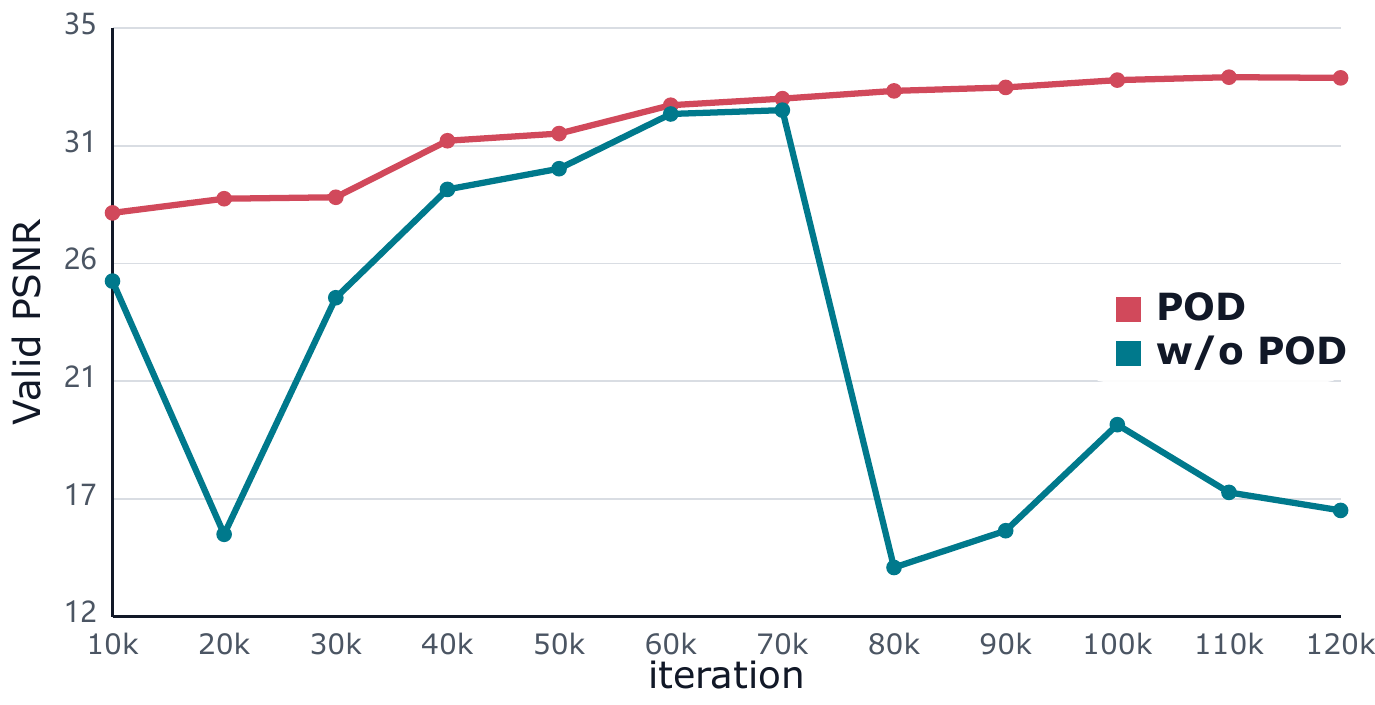}
    \caption{Comparison of log-scale training curves with and without POD. Direct training with image-space rendering loss is highly unstable and can eventually collapse, while POD yields stable optimization.}
    \label{fig:pod}
\end{figure}

The stage-wise residual formulation in Sec.~\ref{sec:stage_wise_residual_gaussian_prediction} decomposes reconstruction into incremental corrections, yet directly training each stage with the residual rendering objective in Eq.~\ref{eq:residual_rendering_objective} remains unstable.
The difficulty arises because the Gaussian parameterization is highly under-constrained with respect to image-space supervision.
As shown in Eq.~\ref{eq:rendering_equation}, the per-pixel color is jointly determined by the center, covariance, and color of each primitive, and these parameters are inherently coupled: changes in spatial extent can be compensated by color magnitude, and shifts in position can be offset by adjusting scale or orientation.
The inverse mapping from an image-space residual to Gaussian parameters is therefore far from unique.
In the multi-stage setting, this ambiguity is further amplified because the target of stage $i$ depends on the reconstruction accumulated by all previous stages.
As shown in Fig.~\ref{fig:pod}, directly learning $\Delta G_i$ from residual rendering supervision leads to highly unstable optimization and can eventually diverge or collapse to degenerate solutions.

Inspired by~\cite{wu2025predict}, we introduce a \emph{Predict--Optimize--Distill} (POD) training strategy.
Instead of relying solely on the residual rendering loss to supervise Gaussian prediction, POD starts from the predicted increments $\Delta G_i=f_{\theta_i}(E_{i-1})$, improves it with a few isolated optimization steps, and then distills the refined result back into the predictor.
In this way, the predictor learns to approximate the result of local iterative Gaussian refinement in a single forward pass, which amortizes the optimization process into the feed-forward network.

\paragraph{Predict.}
At stage $i$, the residual predictor produces additional Gaussian s from $E_{i-1}$, and Stage Control selects the effective subset $\Delta G_i$ as defined in Secs.~\ref{sec:residual_design} and~\ref{sec:stage_control}.
This predicted increments serves as the initialization for the subsequent POD refinement.

\paragraph{Optimize.}
We refine the predicted increments with a few isolated optimization steps.
Starting from a detached trainable copy,
\begin{equation}
    \widehat{\Delta G}_i^{(0)} = \operatorname{copy}(\Delta G_i),
\end{equation}
The copied increments is then updated for $K_i$ steps:
\begin{equation}
    \widehat{\Delta G}_i^{(t+1)}
    =
    \widehat{\Delta G}_i^{(t)}
    -
    \eta
    \nabla_{\widehat{\Delta G}_i^{(t)}}
    \mathcal{L}_{render}^{(i)}
    \left(
        \mathcal{R}
        \left(
            \widehat{\Delta G}_i^{(t)}
        \right),
        E_{i-1}
    \right),
\end{equation}
where $t=0,\dots,K_i-1$, $\eta$ is the learning rate, and $\mathcal{L}_{render}^{(i)}$ denotes the residual rendering objective used for local refinement.\footnote{In practice, the SSIM component of $\mathcal{L}_{render}$ is evaluated on the accumulated rendering $\mathcal{R}(G_{i-1} \cup \widehat{\Delta G}_i^{(t)})$ against $I_{gt}$, since the residual $E_{i-1}$ may contain negative values. Previous-stage Gaussians are kept fixed so that gradients flow only to $\widehat{\Delta G}_i^{(t)}$.}
Further implementation details are provided in the supplementary material.
After $K_i$ steps, we obtain the refined increments
\begin{equation}
    \widehat{\Delta G}_i = \widehat{\Delta G}_i^{(K_i)}.
\end{equation}
This refined increments are then detached and used as the final regression target for distillation.

\paragraph{Distill.}
We use the refined result $\widehat{\Delta G}_i$ as a fixed target to supervise the original prediction $\Delta G_i$ with a Gaussian-space regression loss:
\begin{equation}
    \mathcal{L}_{gaussian}^{(i)}
    =
    \mathcal{L}_{gaussian}\!\left(
        \Delta G_i,\,
        \widehat{\Delta G}_i
    \right).
\end{equation}
Because $\widehat{\Delta G}_i$ is obtained by locally refining the current prediction, it provides a better-aligned target than image-space rendering supervision alone.
This reduces the reliance on unstable image-space gradients and leads to more stable stage-wise training in practice, as shown in Fig.~\ref{fig:pod}.

The overall POD pretraining objective is
\begin{equation}
    \mathcal{L}_{POD}
    =
    \sum_{i=1}^{S}
    \lambda_i \,
    \mathcal{L}_{gaussian}^{(i)},
\end{equation}
where $S$ is the number of stages and $\lambda_i$ balances the contribution of each stage.

POD can be viewed as a distillation-based amortized optimization procedure~\cite{AmortizedOptimization}.
As discussed above, residual rendering supervision alone is under-constrained, and its image-space gradients do not always provide a stable or consistent direction in Gaussian parameter space.
Rather than forcing the predictor to resolve this ambiguity directly, POD mediates the image-space objective through a short optimization followed by distillation.
Starting from the current prediction, the optimizer arrives at a locally refined solution within the same parameter neighborhood.
This refinement preserves one-to-one primitive correspondence with the original prediction, ensuring that the Gaussian-space regression loss used for distillation remains well-defined.
Moreover, the refined solution has been explicitly validated by the residual rendering objective, making it a more reliable on-the-fly supervision target than raw image-space gradients.
By imitating this target, the predictor gradually absorbs the effect of iterative Gaussian refinement into a single feed-forward pass.

\subsection{Finetune and Quantization}
\label{sec:finetune_and_quantization}

\subsubsection{Finetune Training}
\label{sec:finetune_training}

POD training provides a reliable initialization for the stage-wise predictor by distilling locally refined Gaussian targets.
To further improve reconstruction quality without the cost of repeated local refinement, we switch to a lightweight finetuning phase.
During finetuning, the short optimization used in POD is removed, and the entire multi-stage predictor is optimized directly with image-space rendering supervision.
This reduces the per-iteration training cost and enables joint optimization across stages for the final reconstruction.

The stage-wise prediction process remains unchanged during finetuning.
At stage $i$, the network predicts the Gaussian increments $\Delta G_i$ from the residual $E_{i-1}$ with the Stage Control mechanism in Sec.~\ref{sec:stage_control}, producing the accumulated set $G_i = G_{i-1} \cup \Delta G_i$ and its rendering $I_i = \mathcal{R}(G_i)$.

Finetuning removes the short local optimization and the corresponding Gaussian-space distillation targets used in POD.
Instead, each accumulated reconstruction is directly supervised against the ground-truth image:
\begin{equation}
    \mathcal{L}_{\mathrm{ft}}^{(i)}
    =
    \mathcal{L}_{render}(I_i, I_{gt}).
\end{equation}
The overall finetuning objective is
\begin{equation}
    \mathcal{L}_{\mathrm{ft}}
    =
    \sum_{i=1}^{S}\lambda_i \mathcal{L}_{\mathrm{ft}}^{(i)}.
\end{equation}

Unlike POD pretraining, where each stage is supervised by its own refined Gaussian target, finetuning applies image-space supervision to every intermediate accumulated reconstruction $I_i$.
This design serves two purposes: first, because the rendering loss at stage $i$ involves all preceding increments, gradients can propagate back to earlier stages, enabling joint optimization rather than requiring later stages to compensate for upstream errors alone; second, supervising every prefix $G_i$ preserves the progressive nature of AIR, ensuring that each partial reconstruction remains valid on its own while later stages further reduce the residual error.

\subsubsection{Quantization}
\label{sec:quantization}

AIR also supports compact storage of the predicted Gaussian representation.
Following GaussianImage~\cite{gaussianimage}, we apply uniform quantization to the Gaussian parameters.
The detailed bit allocation and decoding rules are provided in the supplementary material.

A key challenge in direct quantization is how to set an appropriate quantization range for each image.
Existing optimization-based pipelines can derive this range from image-specific iterative fitting.
However, this strategy is not suitable for our feed-forward setting, since it would reintroduce per-image post-optimization after prediction.

To avoid this, we combine a global base quantizer with image-adaptive offsets predicted from the ViT~\cite{dosovitskiy2020image} \texttt{[CLS]} token $z_{\mathrm{cls}}$.
For each Gaussian parameter, we maintain a learnable base scale $\bar{\alpha}$ and shift $\bar{\beta}$, and predict bounded image-dependent offsets:
\begin{equation}
    [\Delta \alpha,\, \Delta \beta] = g(z_{\mathrm{cls}}).
\end{equation}
The effective quantization range is then parameterized by
\begin{equation}
    \alpha = \bar{\alpha} + \Delta \alpha,
    \qquad
    \beta = \bar{\beta} + \Delta \beta,
\end{equation}
allowing the quantizer to adapt to each image without iterative range calibration.

During training, we adopt the straight-through estimator (STE)~\cite{bengio2013estimating} for gradients through the non-differentiable rounding operator.
In the finetuning phase, quantization is enabled only on the final-stage Gaussian set $G_S$.
We keep the original stage-wise reconstruction supervision and add a quantization-aware rendering loss on the quantized output:
\begin{equation}
    \mathcal{L}_{q}
    =
    \mathcal{L}_{render}\!\left(
        \mathcal{R}(Q(G_S)),\,
        I_{gt}
    \right)
    +
    \gamma \mathcal{L}_{err},
\end{equation}
where $Q(\cdot)$ denotes the quantizer and $\mathcal{L}_{err}$ penalizes the quantization error.
This trains both the predictor and the quantizer to account for the distortion introduced by quantization.
At inference, the entire pipeline remains feed-forward, while the quantization range is adjusted adaptively for each image.
\section{Experiments}
\label{sec:experiments}

\subsection{Experimental Setup}
\label{sec:implementation}

\textbf{Datasets and Evaluation Metrics.}
We first pretrain AIR on ImageNet~\cite{deng2009imagenet} and then finetune it on DIV2K~\cite{agustsson2017ntire}. 
ImageNet provides large-scale natural image diversity for learning a general feed-forward mapping from image residuals to Gaussian primitives, while DIV2K provides high-resolution images for adapting the model to fine-grained reconstruction. 
We evaluate reconstruction performance on the DIV2K validation set and the Kodak dataset.

We report standard full-reference image quality metrics, including PSNR and MS-SSIM~\cite{wang2003multiscale}, and use LPIPS~\cite{zhang2018unreasonable} with the VGG~\cite{simonyan2014very} backbone to measure perceptual distortion. 
To evaluate representation compactness and efficiency, we also report the number of activated Gaussian primitives and the reconstruction time.

\textbf{Implementation.}
All stages adopt the same predictor architecture.
Given the residual image $E_{i-1}$, a DINOv3 ViT-Base encoder~\cite{simeoni2025dinov3} extracts multi-level features from both shallow and deep transformer layers.
Specifically, we take intermediate features from the 5th and 11th transformer layers and project the encoder output to a 1024-dimensional representation.
A lightweight convolutional head, inspired by MoGe~\cite{wang2025moge,wang2025moge2}, fuses these features into a 32-channel spatial feature map.

We then apply a patch embedding module with patch size $p$.
It partitions the spatial feature map into local patches and maps each patch to a 64-dimensional token.
Each token predicts a Gaussian candidate, forming the dense candidate set $\widetilde{\Delta G}_i$.
The Stage Control mask is then applied to obtain the effective increments $\Delta G_i$.
Therefore, the patch size $p$ directly controls the spatial granularity and the maximum number of candidate Gaussians at each stage.
Smaller patches provide denser prediction sites, while larger patches lead to a more compact representation.
We evaluate several settings, including patch sizes of 5, 6, 7, and 14.

For image-space rendering supervision, we follow Instant-GI and use a weighted combination of pixel-wise distortion and structural similarity:
\begin{equation}
    \mathcal{L}_{render}
    =
    0.7\,\mathcal{L}_1
    +
    0.3\,(1-\mathrm{SSIM}).
\end{equation}
We set the stage weights $\lambda_i$ uniformly across stages.
In POD pretraining, the Gaussian-space distillation loss is multiplied by a coefficient of 100 to match the gradient scale of Gaussian parameter regression.
During finetuning, the image-space reconstruction loss is used with a coefficient of 1.
Further implementation details are provided in the supplementary material.

\textbf{Training Details.}
The encoder is initialized with pretrained DINOv3 weights.
We first train the patch-size-7 model on ImageNet with standard image augmentations.
The model is optimized using AdamW~\cite{loshchilov2017decoupled} with a warmup schedule followed by StepLR decay.

We begin with POD pretraining, which is used to stabilize multi-stage residual Gaussian prediction.
This phase is conducted for 120k iterations with a batch size of 20.
The prediction stages are progressively activated at milestones $[0,20k,40k,60k]$.
When a new stage is activated, its parameters are initialized by copying those from the previous stage, providing a stable starting point for predicting the next residual Gaussian increments.

After POD pretraining, we conduct finetuning and quantization training for 300k iterations in total, with a batch size of 16.
The first 100k iterations perform finetuning without quantization, and the following 200k iterations enable quantization only on the final accumulated Gaussian set $G_S$ as described in Sec.~\ref{sec:quantization}.

Starting from the patch-size-7 checkpoint, we adapt the model to other patch-size settings, including 5, 6, and 14.
These patch-size-specific finetuning runs are conducted for 100k iterations with a batch size of 20.
This adaptation changes the prediction granularity while reusing the pretrained representation, avoiding retraining the full model from scratch. To better handle high-resolution images, we further finetune the model on the DIV2K training set. More training details are provided in the supplementary material.
\subsection{Main Results}

\begin{table}[!t]
    \caption{Quantitative comparison on Kodak with different numbers of Gaussians.}
    \label{tab:kodak_compact}
    \centering
    \resizebox{\linewidth}{!}{
    \begin{tabular}{c c | cccc}
    \toprule
    Method & \#Gaussians & PSNR$\uparrow$ & MS-SSIM$\uparrow$ & LPIPS$\downarrow$ & Time (ms)$\downarrow$ \\
    \midrule
    \multirow{3}{*}{GaussianImage}
    & 28k & 26.98 & 0.914 & 0.401 & 1471 \\
    & 37k & 29.20 & 0.947 & 0.317 & 1463 \\
    & 52k & 30.90 & 0.965 & 0.253 & 1499 \\
    \midrule
    \multirow{3}{*}{Image-GS}
    & 28k & 30.79 & 0.974 & 0.221 & 2494 \\
    & 37k & 31.29 & 0.978 & 0.201 & 2520 \\
    & 52k & 31.90 & 0.982 & 0.179 & 2749 \\
    \midrule
    \multirow{2}{*}{Instant-GI}
    & 25k & 28.52 & 0.948 & 0.315 & 191 \\
    & 43k & 29.67 & 0.957 & 0.273 & 261 \\
    \midrule
    \multirow{3}{*}{\textbf{Ours}}
    & 28k & 30.93 & 0.978 & 0.209 & 157 \\
    & 37k & 31.47 & 0.982 & 0.184 & 159 \\
    & 52k & 32.17 & 0.985 & 0.158 & 162 \\
    \bottomrule
    \end{tabular}
    }
    \end{table}
\begin{table}[t]
    \centering
    \caption{Quantitative comparison on DIV2K with different numbers of Gaussians.}
    \label{tab:div2k_compact}
    \resizebox{\linewidth}{!}{
    \begin{tabular}{c c | cccc}
    \toprule
    Method & \#Gaussians & PSNR$\uparrow$ & MS-SSIM$\uparrow$ & LPIPS$\downarrow$ & Time (ms)$\downarrow$ \\
    \midrule
    \multirow{3}{*}{GaussianImage}
    & 46k & 26.90 & 0.908 & 0.373 & 1879 \\
    & 61k & 28.98 & 0.939 & 0.310 & 1955 \\
    & 86k & 30.77 & 0.958 & 0.258 & 2047 \\
    \midrule
    \multirow{3}{*}{ImageGS}
    & 46k & 30.31 & 0.981 & 0.187 & 3144 \\
    & 61k & 30.80 & 0.984 & 0.169 & 3254 \\
    & 86k & 31.36 & 0.986 & 0.149 & 3323 \\
    \midrule
    \multirow{2}{*}{Instant-GI}
    & 45k & 28.78 & 0.964 & 0.263 & 343 \\
    & 77k & 29.66 & 0.970 & 0.229 & 503 \\
    \midrule
    \multirow{3}{*}{\textbf{Ours}}
    & 46k & 30.76 & 0.984 & 0.193 & 292 \\
    & 61k & 31.30 & 0.987 & 0.171 & 294 \\
    & 86k & 32.02 & 0.989 & 0.145 & 300 \\
    \bottomrule
    \end{tabular}
    }
    \end{table}

\begin{figure*}[t!]
    \centering
    \includegraphics[width=\linewidth]{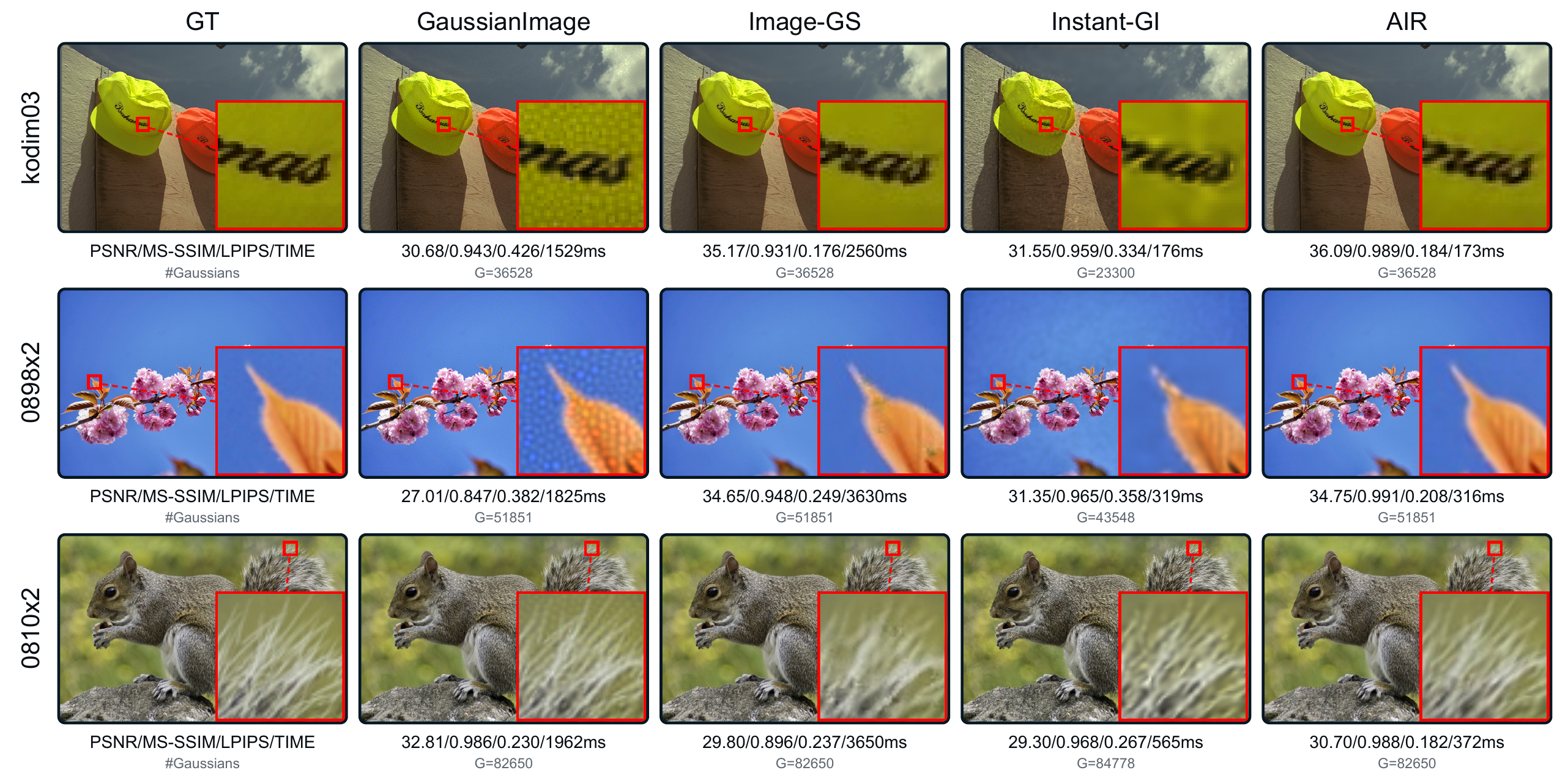}
    \caption{
    Qualitative comparison on Kodak and DIV2K.
    AIR preserves structural details and perceptual quality with a single feed-forward prediction.
    }
    \label{fig:data_vis}
\end{figure*}

Since AIR targets the explicit 2D Gaussian rendering paradigm, we compare with representative methods in this category, including GaussianImage~\cite{gaussianimage}, Image-GS~\cite{ImageGS}, and Instant-GI~\cite{Instant-GI}.
For GaussianImage and Image-GS, we use the same number of Gaussians as AIR.
Since Instant-GI adaptively determines its primitive allocation, we report its original number of Gaussians without forcing it to match ours.

GaussianImage and Image-GS are per-image optimization methods that can yield higher reconstruction quality when given sufficient optimization iterations, as reported in their original papers.
However, such extended optimization incurs substantial encoding time, making it impractical for time-sensitive applications.
To provide a fair comparison under realistic time constraints, we limit GaussianImage to 800 iterations on Kodak and 1000 iterations on DIV2K, and Image-GS to 600 iterations on both datasets with an additional densification step after 300 iterations.
For Instant-GI, we report the quality of its direct network initialization without subsequent refinement.
We emphasize that this evaluation is not intended to undermine the capability of these methods, but rather to assess the reconstruction quality each approach can achieve under practical encoding-time constraints.
Tables~\ref{tab:kodak_compact} and~\ref{tab:div2k_compact} report the quantitative results.

Under this setting, GaussianImage and Image-GS demonstrate that 2D Gaussian primitives can effectively represent images, yet their quality remains bounded by the limited number of optimization iterations.
AIR amortizes the fitting process into a feed-forward predictor and reconstructs images within 160--300 ms, eliminating the need for iterative per-image optimization entirely.
Under the same number of Gaussians, AIR reaches comparable or better reconstruction quality than these optimization-based baselines.

\begin{figure}[t!]
    \centering
    \includegraphics[width=\linewidth]{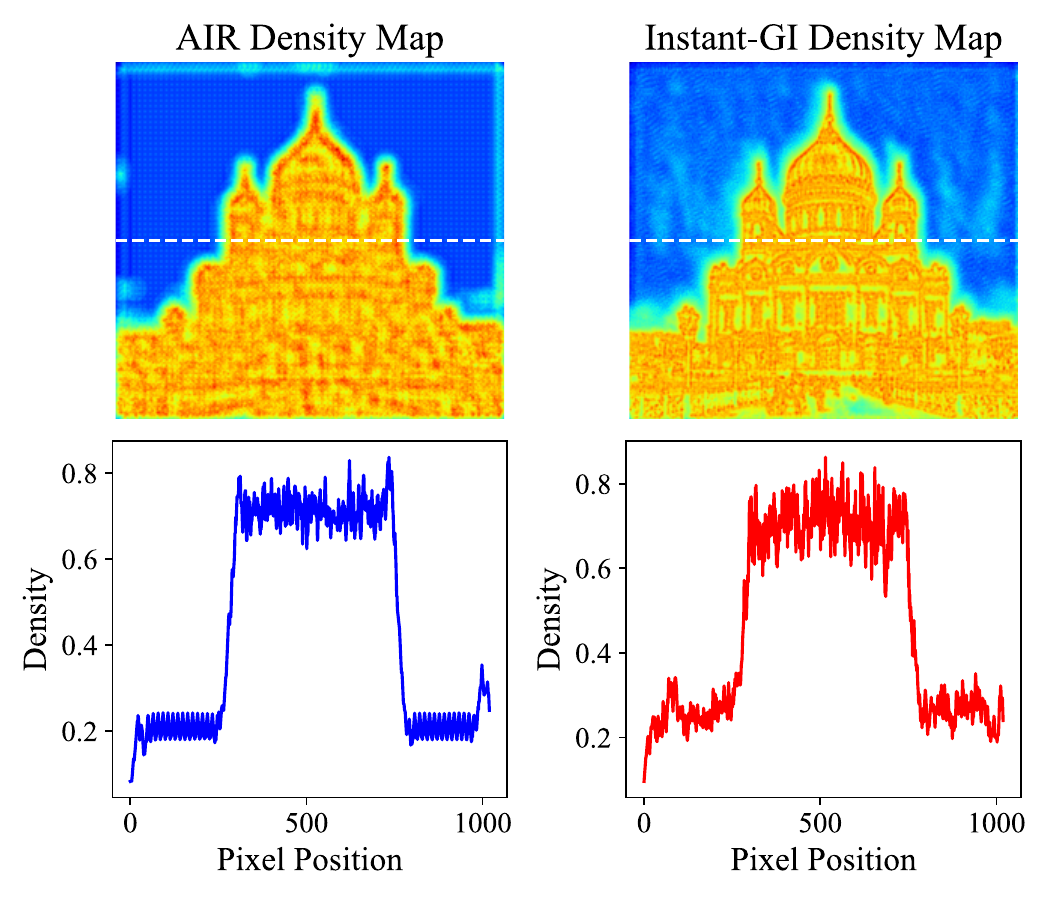}
    \caption{
    Comparison of Gaussian density maps between AIR and Instant-GI.
    AIR produces adaptive, structure-aware Gaussian allocation without PPM pseudo labels.
    The bottom row shows density profiles along the dashed horizontal line.
    }
    \label{fig:density_map}
\end{figure}

The most direct comparison is with Instant-GI, since both methods use network prediction to avoid long per-image optimization.
Instant-GI provides fast initialization with an adaptive number of Gaussians, but it still relies on handcrafted geometric priors to construct the initial Gaussians.
These priors help organize the primitives, but may also limit direct reconstruction quality and make further refinement necessary.
In contrast, AIR uses stage-wise residual prediction and Stage Control to adaptively activate Gaussians in under-reconstructed regions without prior-driven geometric construction.
As shown in Tables~\ref{tab:kodak_compact} and~\ref{tab:div2k_compact}, AIR consistently outperforms Instant-GI by a significant margin across all metrics on both datasets, demonstrating stronger pixel-level accuracy as well as better structural and perceptual fidelity.

Fig.~\ref{fig:density_map} further compares the Gaussian density distributions of the two methods.
Instant-GI achieves adaptive primitive allocation through an explicit position probability map (PPM) followed by dithering.
Although AIR does not use PPM pseudo labels, it still produces a similarly structure-aware density distribution: more Gaussians are allocated to complex foreground regions, while fewer are assigned to smooth background areas.
The density profiles along the sampled scanline show that AIR responds to image structure in a manner comparable to Instant-GI.
This suggests that adaptive Gaussian allocation can be learned directly from reconstruction supervision, without relying on PPM pseudo labels that require lengthy pre-training to construct and inevitably encode the inductive bias of their underlying density estimator.

The qualitative results in Fig.~\ref{fig:data_vis} are consistent with the quantitative trends in Tab.~\ref{tab:kodak_compact} and Tab.~\ref{tab:div2k_compact}.
Optimization-based baselines may still show under-optimized details under limited iterations, while Instant-GI tends to miss fine structures in challenging regions.
AIR produces more faithful feed-forward reconstructions, achieving a favorable balance among reconstruction quality, Gaussian compactness, and runtime efficiency.
\subsection{Ablation Study}
\label{sec:ablation}

We conduct ablation studies to evaluate the effect of the proposed design and training strategy.

\subsubsection{One-shot vs. Progressive Stage-wise Prediction}
\begin{table}[!t]
    \caption{Comparison between one-shot and progressive stage-wise prediction.}
    \label{tab:ablation_one_shot_stagewise}
    \centering
    \resizebox{\linewidth}{!}{
    \begin{tabular}{c c | ccccc}
    \toprule
    Prediction Mode & Setting & PSNR$\uparrow$ & MS-SSIM$\uparrow$ & LPIPS$\downarrow$ & \#Gaussians & Time (ms)$\downarrow$ \\
    \midrule
    \begin{tabular}[c]{@{}c@{}}One-shot\\($ps=7$)\end{tabular} & stage=1 & 26.58 & 0.950 & 0.336 & 14528 & 74 \\
    \midrule
    \multirow{4}{*}{\begin{tabular}[c]{@{}c@{}}Stage-wise\\($ps=14$)\end{tabular}}
    & stage=1 & 21.91 & 0.823 & 0.512 & 3658  & 73 \\
    & stage=2 & 23.91 & 0.897 & 0.440 & 6808  & 146 \\
    & stage=3 & 25.25 & 0.926 & 0.385 & 9819  & 219 \\
    & stage=4 & 26.54 & 0.947 & 0.348 & 12690 & 293 \\
    \bottomrule
    \end{tabular}
    }
\end{table}

We compare the proposed progressive stage-wise predictor with a one-shot baseline on DIV2K under a comparable number of Gaussians.
The one-shot baseline uses a patch size of 7 and predicts the Gaussian representation in a single stage.
We report its POD-pretrained checkpoint, as it provides a stronger standalone Stage-1 result than the finetuned checkpoint.
The stage-wise model uses a patch size of 14 and progressively accumulates Gaussian primitives, reaching a similar number of Gaussians at the final stage.

As shown in Tab.~\ref{tab:ablation_one_shot_stagewise}, the stage-wise predictor provides a controllable trade-off between reconstruction quality and representation compactness.
As more stages are activated, the reconstruction quality improves consistently, with PSNR increasing from 21.91 dB to 26.54 dB from Stage-1 to Stage-4.
Meanwhile, the number of Gaussians increases progressively rather than being allocated all at once.
This trend indicates that later stages effectively compensate for the residual errors left by previous stages.

Compared with the one-shot baseline, the final stage-wise result achieves comparable reconstruction quality, with 26.54 dB PSNR versus 26.58 dB, while using fewer Gaussians, 12690 versus 14528.
Although the residual stage-wise design does not clearly improve the final reconstruction quality over one-shot prediction, it enables explicit control over the number of activated Gaussians.
This allows AIR to adjust the representation size progressively according to the desired quality--compactness trade-off.

\begin{figure}[t!]
    \centering
    \includegraphics[width=\linewidth]{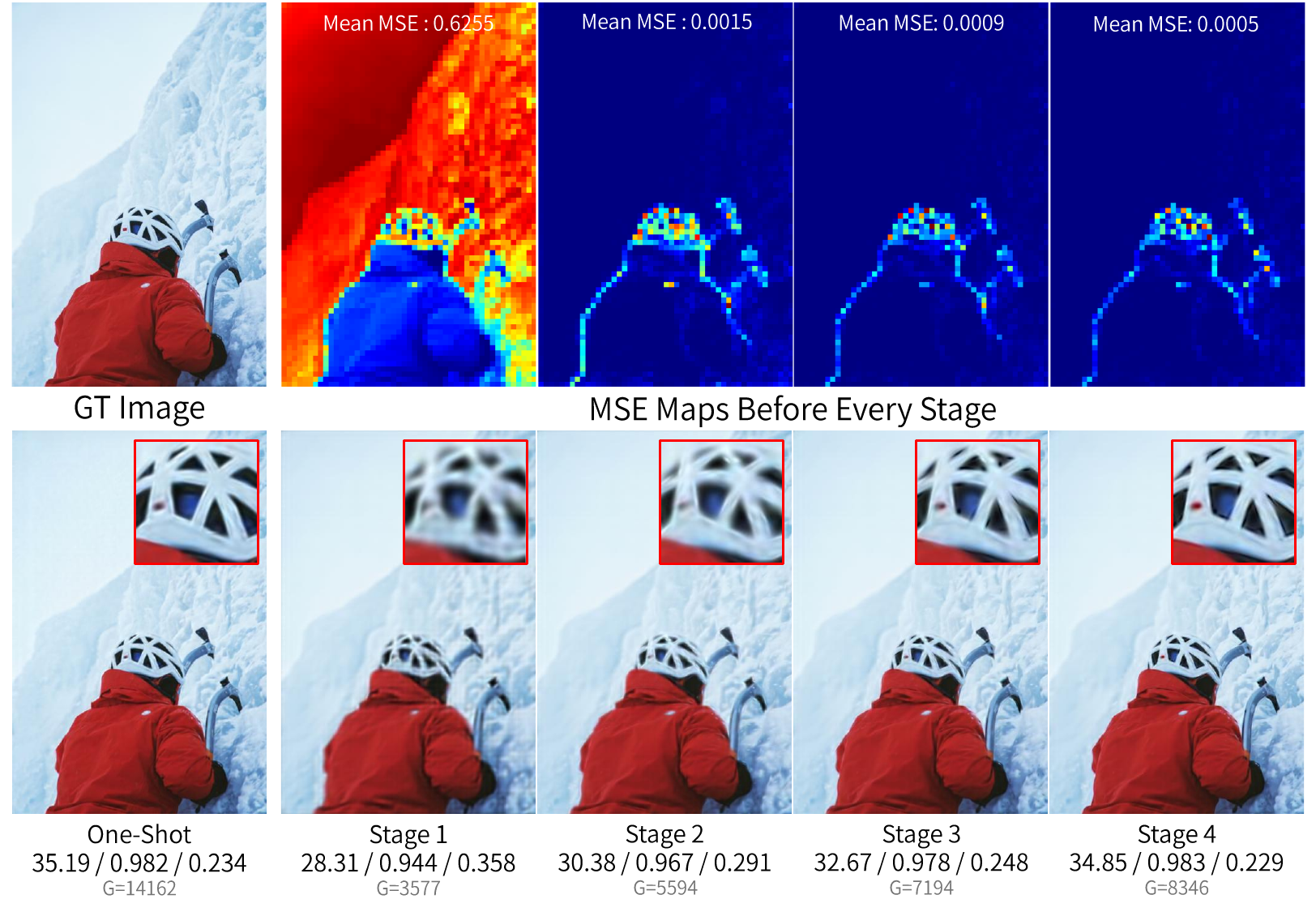}
    \caption{
        Visualization of progressive stage-wise prediction.
        The top row shows the ground-truth image and the pre-stage MSE maps before each stage.
        The bottom row compares the one-shot reconstruction with intermediate reconstructions produced by different stages.
    }
    \label{fig:stage_vis}
\end{figure}

Fig.~\ref{fig:stage_vis} visualizes the progressive behavior of the proposed stage-wise predictor.
As more stages are activated, the reconstruction is gradually refined, and the pre-stage MSE maps show that residual errors decrease consistently.
The remaining errors become increasingly concentrated around high-frequency structures and object boundaries, indicating that later stages mainly focus on under-reconstructed regions.
Although the final stagewise result does not clearly outperform the one-shot baseline in this example, it provides a controllable, progressive reconstruction process by gradually increasing the number of activated Gaussians, achieving a similar visual performance with fewer Gaussians.

\subsubsection{Stage Control Strategy}
\begin{table}[!t]
    \caption{Ablation of stage-control activation criteria.}
    \label{tab:ablation_stage_control}
    \centering
    \resizebox{\linewidth}{!}{
    \begin{tabular}{c c | ccccc}
    \toprule
    PSNR Target & SSIM Target & PSNR$\uparrow$ & MS-SSIM$\uparrow$ & LPIPS$\downarrow$ & \#Gaussians & Utilization$\downarrow$ \\
    \midrule
    -- & -- & 30.89 & 0.985 & 0.169 & 58114 & 1.00 \\
    35 dB & -- & 30.60 & 0.982 & 0.214 & 40163 & 0.69 \\
    -- & 0.95 & 30.29 & 0.983 & 0.199 & 43750 & 0.75 \\
    \textbf{35 dB} & \textbf{0.95} & 30.76 & 0.984 & 0.193 & 46001 & 0.79 \\
    40 dB & 0.98 & 30.87 & 0.985 & 0.177 & 52318 & 0.90 \\
    \bottomrule
    \end{tabular}
    }
    \end{table}

\begin{figure}[t!]
    \centering
    \includegraphics[width=\linewidth]{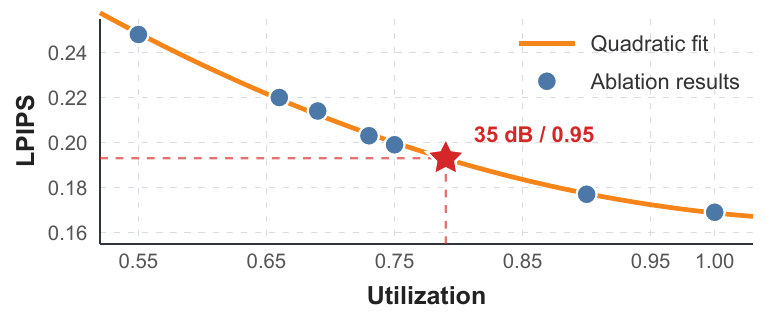}
    \caption{Gaussian utilization and LPIPS under different PSNR/SSIM stage-control thresholds.}
    \label{fig:utilization_lpips}
\end{figure}

We further ablate the proposed stage control strategy.
All variants are evaluated under the same training and inference setting, and differ only in the activation criteria used for stage control.
Here, ``-'' denotes that no filtering mask is applied.
The utilization ratio measures the proportion of activated Gaussians among all candidate Gaussians, i.e., $4 \times H/p \times W/p$.

As shown in Tab.~\ref{tab:ablation_stage_control}, the no-mask setting gives the best reconstruction quality by activating all candidate Gaussians, but also uses the most primitives.
In contrast, stage control follows a quality-constrained allocation principle: it activates new Gaussians only where the current reconstruction fails to meet the prescribed fidelity targets, thereby reducing unnecessary primitives under a controlled distortion level.

The PSNR and SSIM targets provide complementary cues for stage control.
The PSNR target detects regions with large local reconstruction errors, while the SSIM target captures structural distortions that are not fully reflected by pixel-wise error.
Using either target alone therefore gives an incomplete quality criterion.
Combining them allows the activation mask to cover both pixel-level errors and structural degradation, which explains why the joint setting improves over the MSE-only and SSIM-only variants in Tab.~\ref{tab:ablation_stage_control}.

Fig.~\ref{fig:utilization_lpips} further explains our choice of thresholds.
As the thresholds become stricter, more Gaussians are activated and better LPIPS is achieved.
However, the perceptual gain becomes smaller when utilization continues to increase.
The operating point of 35~dB PSNR and 0.95 SSIM keeps the utilization at a moderate level while maintaining competitive perceptual quality, making it a reasonable default choice for stage control.

\subsubsection{Training Strategy}

To validate the effectiveness of the proposed training strategy, we analyze POD pretraining and the subsequent finetuning phase.
POD pretraining first stabilizes multi-stage residual Gaussian prediction and prevents training collapse.
The finetuning phase then builds on this stable initialization and further improves reconstruction quality through cross-stage coordination under image-space supervision.

As shown in Table~\ref{tab:ablation_pod_stability} and Fig.~\ref{fig:pod}, direct supervision with the rendering loss is not completely ineffective.
Direct Supervision reaches a best PSNR of 32.13 dB during training, indicating that the mapping from residual images to Gaussian parameters can be partially learned in the short term.
However, its final PSNR drops sharply to 16.22 dB, and the training collapses at iteration 80k.
This result suggests that direct supervision can temporarily find a reasonable solution, but the training process is unstable.
This instability is consistent with the analysis in Sec.~\ref{sec:pod_training}.
Direct image-space supervision faces a non-unique inverse mapping from residual images to Gaussian parameters, and this ambiguity is amplified across stages through error accumulation and parameter compensation.

\begin{table}[!t]
    \caption{Ablation of POD for training stability.}
    \label{tab:ablation_pod_stability}
    \centering
    \resizebox{\linewidth}{!}{
    \begin{tabular}{c | ccc}
    \toprule
    Training Strategy & Best PSNR$\uparrow$ & Final PSNR$\uparrow$ & Collapse Iteration \\
    \midrule
    Direct Supervision & 32.13 & 16.22 & 80k \\
    \textbf{POD} & 33.44 & 33.41 & N/A \\
    \bottomrule
    \end{tabular}
    }
    \end{table}

In contrast, Full POD reaches a higher best PSNR of 33.44 dB and remains stable, with a final PSNR of 33.41 dB and no collapse.
This indicates that POD mainly improves training stability rather than only increasing the peak reconstruction quality.
By distilling the short-horizon optimized Gaussian increments, POD provides a more reliable regression target for the predictor and reduces the amortization gap between feed-forward prediction and iterative fitting.

\begin{figure}[t!]
    \centering
    \includegraphics[width=\linewidth]{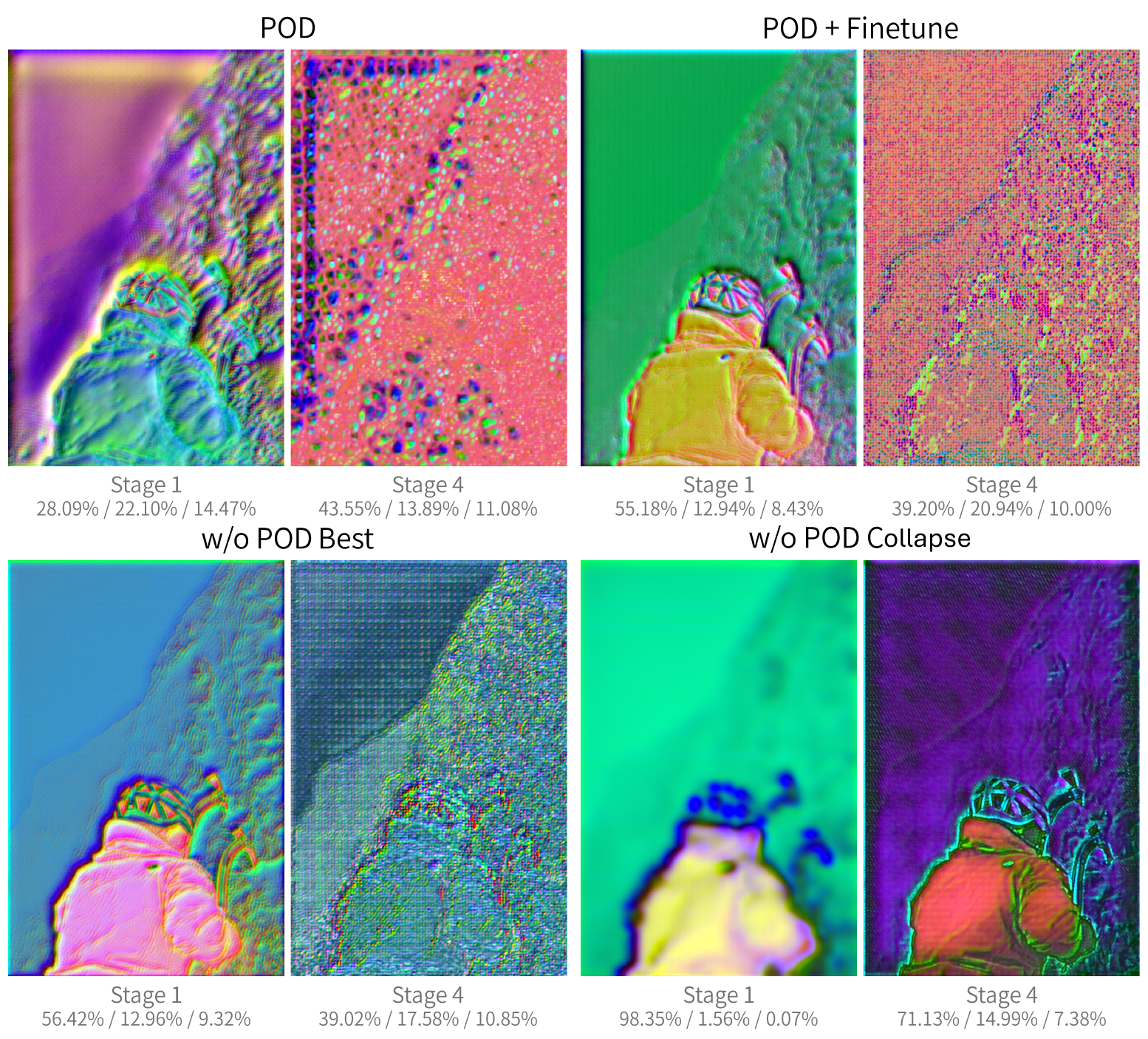}
    \caption{
    PCA visualization of stage features under different training strategies.
    For each feature map, the first three principal components are visualized as RGB channels, and the values indicate their explained variance ratios.
    }
    \label{fig:feat_pca}
\end{figure}

Fig.~\ref{fig:feat_pca} further supports the stabilizing effect of POD at the feature level.
In the \emph{w/o POD Collapse} case, the Stage-1 feature is almost dominated by a single principal component, which explains 98.35\% of the variance.
This near one-dimensional feature space is consistent with the training collapse observed in Tab.~\ref{tab:ablation_pod_stability}.

By contrast, POD produces a more balanced feature distribution, with the first three principal components explaining 28.09\%, 22.10\%, and 14.47\% of the variance.
The corresponding PCA maps retain clearer spatial structures, indicating that POD helps preserve informative features for residual Gaussian prediction.
This supports the claim that POD stabilizes training not only at the reconstruction level, but also at the representation level.

\begin{table}[!t]
    \caption{Ablation of the finetuning phase.}
    \label{tab:ablation_finetune_stagewise}
    \centering
    \resizebox{\linewidth}{!}{
            \begin{tabular}{c | cccc}
            \toprule
            Training Strategy & After Stage 1$\uparrow$ & After Stage 2$\uparrow$ & After Stage 3$\uparrow$ & After Stage 4$\uparrow$ \\
            \midrule
            Only POD & 26.58 & 28.29 & 29.07 & 29.48 \\
            \textbf{POD+finetune} & 25.58 & 27.89 & 29.52 & 30.76 \\
            \midrule
            $\Delta$ & -1.00 & -0.40 & +0.45 & +1.28 \\
            \bottomrule
            \end{tabular}
    }
\end{table}

After POD pretraining, we further examine the effect of the finetuning phase.
As shown in Tab.~\ref{tab:ablation_finetune_stagewise}, POD+finetune slightly lowers the cumulative PSNR after the first two stages, but improves the later-stage results, especially the final Stage-4 reconstruction.
This indicates that finetuning does not simply strengthen each stage independently.
Instead, it redistributes the reconstruction burden across stages and improves the final multi-stage output.

This behavior reflects the difference between stage-wise local fitting and cross-stage joint optimization.
With Only POD, each stage is trained to match its own optimized residual target, which favors strong intermediate reconstructions but limits coordination across stages.
During the finetuning phase, the model is supervised by the cumulative rendering result, allowing the reconstruction burden to be redistributed across stages.
Therefore, early stages may sacrifice some standalone PSNR to provide a better basis for later residual correction.
This leads to lower PSNR after Stage-1 and Stage-2, but higher final reconstruction quality after Stage-4.

This trend is also reflected in the PCA visualization~\cite{abdi2010principal} in Fig.~\ref{fig:feat_pca}.
After finetuning, the Stage-1 feature shows clearer structural patterns, while the Stage-4 feature contains more high-frequency residual responses.
This suggests that finetuning encourages different stages to take more specialized roles for the final accumulated reconstruction.

Overall, POD and the finetuning phase address two successive training issues.
POD stabilizes residual Gaussian prediction and prevents collapse, while the finetuning phase promotes cross-stage coordination and improves the final accumulated reconstruction quality.

\subsubsection{Quantization}

We evaluate the proposed quantization scheme on DIV2K with a patch size of 7.
We compare with GaussianImage and Image-GS under a comparable number of Gaussians.
GaussianImage follows its original quantization settings and applies 300 iterations of refinement on the converged full-precision representation.
Image-GS is optimized from scratch for 600 iterations, with per-attribute bit-widths of 16 for position and scale and 8 for rotation and color.
Since Instant-GI does not provide a feed-forward quantization mechanism and its compressed representation still depends on subsequent GaussianImage-style refinement, it is excluded from this comparison.
For AIR, we further compare three quantization variants with the same bit allocation:
\emph{Per-image Range}, which directly uses the parameter range of each predicted Gaussian set;
\emph{Learned Global Range}, which learns a fixed global quantization range for all images;
and the proposed \emph{Adaptive Range}, which further adjusts the global range with image-dependent offsets.

\begin{table}[!t]
    \caption{Quantization ablation and comparison on DIV2K.}  
    \label{tab:ablation_quantizer}
    \centering
    \resizebox{\linewidth}{!}{
    \begin{tabular}{c | ccccc}
    \toprule
    Quantizer & PSNR$\uparrow$ & MS-SSIM$\uparrow$ & LPIPS$\downarrow$ & bpp$\downarrow$ & Time(ms)$\downarrow$ \\
    \midrule
    GaussianImage                      & 26.42 & 0.900 & 0.391 & 3.51 & 4446 \\
    Image-GS                           & 30.17 & 0.979 & 0.193 & 6.04 & 3091 \\
    \midrule
    AIR + Per-image Range              & 29.37 & 0.973 & 0.234 & 3.14 & 298 \\
    AIR + Learned Global Range          & 29.17 & 0.975 & 0.223 & 3.21 & 298 \\
    \textbf{AIR + Adaptive Range}       & 30.23 & 0.979 & 0.214 & 3.28 & 300 \\
    AIR w/o Quantization               & 30.76 & 0.984 & 0.193 & --   & 292 \\
    \bottomrule
    \end{tabular}
    }
\end{table}

As shown in Table~\ref{tab:ablation_quantizer}, AIR with Adaptive Range achieves competitive reconstruction quality with substantially lower reconstruction time compared with GaussianImage and Image-GS.
It reaches 30.23~dB PSNR, comparable to Image-GS (30.17~dB), while using a significantly lower bitrate (3.28 vs.\ 6.04~bpp) and avoiding per-image iterative fitting.
Instead of optimizing a full-precision Gaussian representation before compression, AIR directly quantizes the feed-forward predicted Gaussian set, preserving the fully feed-forward pipeline.

We further compare different AIR quantization strategies.
Per-image Range achieves slightly higher PSNR than Learned Global Range, suggesting that image-dependent parameter distributions should be considered during quantization.
However, directly using the raw per-image range can be affected by outliers in the predicted parameters and is not optimized for final reconstruction quality.
Learned Global Range introduces learnable quantization parameters, but a single fixed range is insufficient to handle both cross-dataset domain shift and image-level distribution variations.

In contrast, the proposed Adaptive Range combines a learned global range with image-dependent offsets.
It improves PSNR from 29.17~dB to 30.23~dB over Learned Global Range, with consistent improvements in MS-SSIM and LPIPS.
Compared with the unquantized result (30.76~dB), Adaptive Range introduces only a 0.53~dB drop, indicating that the feed-forward predicted Gaussians are amenable to direct parameter quantization.
Although Adaptive Range slightly increases bpp relative to the other two variants, this is consistent with the entropy-coding behavior that a less concentrated distribution of quantized symbols may require a higher bitrate.
This small bitrate increase is therefore a reasonable trade-off for preserving finer Gaussian attribute variations and reducing quantization-induced distortion.

\section{Conclusion}

In this paper, we presented AIR, a self-supervised feed-forward framework for amortized 2D Gaussian image reconstruction. 
AIR replaces costly per-image Gaussian optimization with direct network prediction, enabling efficient reconstruction without test-time iterative fitting. 
To support adaptive primitive allocation, AIR introduces a stage-wise residual prediction framework with explicit stage control, where new Gaussians are progressively added only to under-reconstructed regions. 
Predict--Optimize--Distill training stabilizes each residual prediction stage, while subsequent multi-stage finetuning improves the final accumulated reconstruction. 
AIR also incorporates a lightweight adaptive quantization module for compact storage of the predicted Gaussian primitives.
Experiments on Kodak and DIV2K show that AIR achieves high-quality reconstruction with short feed-forward inference time. 
Compared with optimization-based Gaussian reconstruction methods, AIR avoids iterative fitting; compared with Instant-GI, it improves direct reconstruction quality without PPM pseudo labels or handcrafted priors.

A current limitation is that different stages use separate model parameters, which introduces redundancy and leaves room for more stage-agnostic architectures. 
In addition, our quantization module adopts a relatively simple uniform quantization scheme, and more advanced entropy coding or learned compression strategies could further improve the storage efficiency of the predicted Gaussian representation.

\section*{Acknowledgment}
This work is supported by the Natural Science Foundation of China under Grant 62302174. The computation is completed in the HPC Platform of Huazhong University of Science and Technology. We also thank Farsee2 Technology Ltd for providing devices to support the validation of our method.


\bibliographystyle{IEEEtran}
\bibliography{references}

@inproceedings{gaussianimage,
  title={Gaussianimage: 1000 fps image representation and compression by 2d gaussian splatting},
  author={Zhang, Xinjie and Ge, Xingtong and Xu, Tongda and He, Dailan and Wang, Yan and Qin, Hongwei and Lu, Guo and Geng, Jing and Zhang, Jun},
  booktitle={European Conference on Computer Vision},
  pages={327--345},
  year={2024},
  organization={Springer}
}

@inproceedings{ImageGS,
  title={Image-gs: Content-adaptive image representation via 2d gaussians},
  author={Zhang, Yunxiang and Li, Bingxuan and Kuznetsov, Alexandr and Jindal, Akshay and Diolatzis, Stavros and Chen, Kenneth and Sochenov, Anton and Kaplanyan, Anton and Sun, Qi},
  booktitle={Proceedings of the Special Interest Group on Computer Graphics and Interactive Techniques Conference Conference Papers},
  pages={1--11},
  year={2025}
}

@InProceedings{Instant-GI,
    author    = {Zeng, Zhaojie and Wang, Yuesong and Guan, Tao and Yang, Chao and Ju, Lili},
    title     = {Instant GaussianImage: A Generalizable and Self-Adaptive Image Representation via 2D Gaussian Splatting},
    booktitle = {Proceedings of the IEEE/CVF International Conference on Computer Vision (ICCV)},
    month     = {October},
    year      = {2025},
    pages     = {27896-27905}
}

@article{AmortizedOptimization,
  title={Tutorial on amortized optimization},
  author={Amos, Brandon},
  journal={Foundations and Trends in Machine Learning},
  volume={16},
  number={5},
  pages={592--732},
  year={2023},
  publisher={Emerald Publishing Limited}
}

@article{wang2004image,
  title={Image quality assessment: from error visibility to structural similarity},
  author={Wang, Zhou and Bovik, Alan C and Sheikh, Hamid R and Simoncelli, Eero P},
  journal={IEEE transactions on image processing},
  volume={13},
  number={4},
  pages={600--612},
  year={2004},
  publisher={IEEE}
}

@article{NeRF,
  title={Nerf: Representing scenes as neural radiance fields for view synthesis},
  author={Mildenhall, Ben and Srinivasan, Pratul P and Tancik, Matthew and Barron, Jonathan T and Ramamoorthi, Ravi and Ng, Ren},
  journal={Communications of the ACM},
  volume={65},
  number={1},
  pages={99--106},
  year={2021},
  publisher={ACM New York, NY, USA}
}

@inproceedings{chen2021learning,
  title={Learning continuous image representation with local implicit image function},
  author={Chen, Yinbo and Liu, Sifei and Wang, Xiaolong},
  booktitle={Proceedings of the IEEE/CVF conference on computer vision and pattern recognition},
  pages={8628--8638},
  year={2021}
}

@inproceedings{sitzmann2020siren,
  title     = {Implicit Neural Representations with Periodic Activation Functions},
  author    = {Sitzmann, Vincent and Martel, Julien N. P. and Bergman, Alexander W. and Lindell, David B. and Wetzstein, Gordon},
  booktitle = {Advances in Neural Information Processing Systems},
  volume    = {33},
  pages     = {7462--7473},
  year      = {2020}
}

@article{dupont2021coin,
  title         = {{COIN}: {CO}mpression with Implicit Neural Representations},
  author        = {Dupont, Emilien and Goli{\'n}ski, Adam and Alizadeh, Milad and Teh, Yee Whye and Doucet, Arnaud},
  journal       = {arXiv preprint arXiv:2103.03123},
  year          = {2021},
  eprint        = {2103.03123},
  archivePrefix = {arXiv},
  primaryClass  = {cs.CV}
}

@article{dupont2022coinpp,
  title   = {{COIN++}: Neural Compression Across Modalities},
  author  = {Dupont, Emilien and Loya, Hrushikesh and Alizadeh, Milad and Goli{\'n}ski, Adam and Teh, Yee Whye and Doucet, Arnaud},
  journal = {Transactions on Machine Learning Research},
  year    = {2022},
  url     = {https://openreview.net/forum?id=NXB0rEM2Tq}
}

@article{muller2022instantngp,
  title   = {Instant Neural Graphics Primitives with a Multiresolution Hash Encoding},
  author  = {M{\"u}ller, Thomas and Evans, Alex and Schied, Christoph and Keller, Alexander},
  journal = {ACM Transactions on Graphics},
  volume  = {41},
  number  = {4},
  pages   = {102:1--102:15},
  year    = {2022},
  doi     = {10.1145/3528223.3530127}
}

@inproceedings{saragadam2023wire,
  title     = {{WIRE}: Wavelet Implicit Neural Representations},
  author    = {Saragadam, Vishwanath and LeJeune, Daniel and Tan, Jasper and Balakrishnan, Guha and Veeraraghavan, Ashok and Baraniuk, Richard G.},
  booktitle = {Proceedings of the IEEE/CVF Conference on Computer Vision and Pattern Recognition},
  pages     = {18507--18516},
  year      = {2023}
}

@inproceedings{kerbl20233dgaussians,
  title     = {3D Gaussian Splatting for Real-Time Radiance Field Rendering},
  author    = {Kerbl, Bernhard and Kopanas, Georgios and Leimk{\"u}hler, Thomas and Drettakis, George},
  booktitle = {ACM SIGGRAPH 2023 Conference Proceedings},
  pages     = {1--14},
  year      = {2023},
  doi       = {10.1145/3588432.3592433}
}

@inproceedings{marino2018iterative,
  title     = {Iterative Amortized Inference},
  author    = {Marino, Joseph and Yue, Yisong and Mandt, Stephan},
  booktitle = {Proceedings of the 35th International Conference on Machine Learning},
  pages     = {3403--3412},
  year      = {2018}
}

@article{adler2018learned,
  title   = {Learned Primal-Dual Reconstruction},
  author  = {Adler, Jonas and {\"O}ktem, Ozan},
  journal = {IEEE Transactions on Medical Imaging},
  volume  = {37},
  number  = {6},
  pages   = {1322--1332},
  year    = {2018},
  doi     = {10.1109/TMI.2018.2799231}
}

@inproceedings{putzky2017recurrent,
  title     = {Recurrent Inference Machines for Solving Inverse Problems},
  author    = {Putzky, Patrick and Welling, Max},
  booktitle = {Workshop Track of the International Conference on Learning Representations},
  year      = {2017}
}

@inproceedings{wu2025predict,
  title     = {Predict-Optimize-Distill: A Self-Improving Cycle for 4D Object Understanding},
  author    = {Wu, Mingxuan and Huang, Huang and Kerr, Justin and Kim, Chung Min and Zhang, Anthony and Yi, Brent and Kanazawa, Angjoo},
  booktitle = {Proceedings of the IEEE/CVF International Conference on Computer Vision},
  pages     = {6575--6584},
  year      = {2025}
}

@inproceedings{wang2025moge,
  title={Moge: Unlocking accurate monocular geometry estimation for open-domain images with optimal training supervision},
  author={Wang, Ruicheng and Xu, Sicheng and Dai, Cassie and Xiang, Jianfeng and Deng, Yu and Tong, Xin and Yang, Jiaolong},
  booktitle={Proceedings of the Computer Vision and Pattern Recognition Conference},
  pages={5261--5271},
  year={2025}
}

@misc{wang2025moge2,
      title={MoGe-2: Accurate Monocular Geometry with Metric Scale and Sharp Details}, 
      author={Ruicheng Wang and Sicheng Xu and Yue Dong and Yu Deng and Jianfeng Xiang and Zelong Lv and Guangzhong Sun and Xin Tong and Jiaolong Yang},
      year={2025},
      eprint={2507.02546},
      archivePrefix={arXiv},
      primaryClass={cs.CV},
      url={https://arxiv.org/abs/2507.02546}, 
}

@inproceedings{klocek2019hypernetwork,
  title={Hypernetwork functional image representation},
  author={Klocek, Sylwester and Maziarka, {\L}ukasz and Wo{\l}czyk, Maciej and Tabor, Jacek and Nowak, Jakub and {\'S}mieja, Marek},
  booktitle={International Conference on Artificial Neural Networks},
  pages={496--510},
  year={2019},
  organization={Springer}
}

@inproceedings{deng2009imagenet,
  title={Imagenet: A large-scale hierarchical image database},
  author={Deng, Jia and Dong, Wei and Socher, Richard and Li, Li-Jia and Li, Kai and Fei-Fei, Li},
  booktitle={2009 IEEE conference on computer vision and pattern recognition},
  pages={248--255},
  year={2009},
  organization={Ieee}
}

@inproceedings{agustsson2017ntire,
  title={Ntire 2017 challenge on single image super-resolution: Dataset and study},
  author={Agustsson, Eirikur and Timofte, Radu},
  booktitle={Proceedings of the IEEE conference on computer vision and pattern recognition workshops},
  pages={126--135},
  year={2017}
}

@inproceedings{zhang2018unreasonable,
  title={The unreasonable effectiveness of deep features as a perceptual metric},
  author={Zhang, Richard and Isola, Phillip and Efros, Alexei A and Shechtman, Eli and Wang, Oliver},
  booktitle={Proceedings of the IEEE conference on computer vision and pattern recognition},
  pages={586--595},
  year={2018}
}

@inproceedings{wang2003multiscale,
  title={Multiscale structural similarity for image quality assessment},
  author={Wang, Zhou and Simoncelli, Eero P and Bovik, Alan C},
  booktitle={The thrity-seventh asilomar conference on signals, systems \& computers, 2003},
  volume={2},
  pages={1398--1402},
  year={2003},
  organization={Ieee}
}

@article{loshchilov2017decoupled,
  title={Decoupled weight decay regularization},
  author={Loshchilov, Ilya and Hutter, Frank},
  journal={arXiv preprint arXiv:1711.05101},
  year={2017}
}

@article{bengio2013estimating,
  title={Estimating or propagating gradients through stochastic neurons for conditional computation},
  author={Bengio, Yoshua and L{\'e}onard, Nicholas and Courville, Aaron},
  journal={arXiv preprint arXiv:1308.3432},
  year={2013}
}

@article{simeoni2025dinov3,
  title={Dinov3},
  author={Sim{\'e}oni, Oriane and Vo, Huy V and Seitzer, Maximilian and Baldassarre, Federico and Oquab, Maxime and Jose, Cijo and Khalidov, Vasil and Szafraniec, Marc and Yi, Seungeun and Ramamonjisoa, Micha{\"e}l and others},
  journal={arXiv preprint arXiv:2508.10104},
  year={2025}
}

@article{xu2025resplat,
  title={ReSplat: Learning Recurrent Gaussian Splatting},
  author={Xu, Haofei and Barath, Daniel and Geiger, Andreas and Pollefeys, Marc},
  journal={arXiv preprint arXiv:2510.08575},
  year={2025}
}

@article{ongie2020deep,
  title={Deep learning techniques for inverse problems in imaging},
  author={Ongie, Gregory and Jalal, Ajil and Metzler, Christopher A and Baraniuk, Richard G and Dimakis, Alexandros G and Willett, Rebecca},
  journal={IEEE Journal on Selected Areas in Information Theory},
  volume={1},
  number={1},
  pages={39--56},
  year={2020},
  publisher={IEEE}
}

@inproceedings{xie2022neural,
  title={Neural fields in visual computing and beyond},
  author={Xie, Yiheng and Takikawa, Towaki and Saito, Shunsuke and Litany, Or and Yan, Shiqin and Khan, Numair and Tombari, Federico and Tompkin, James and Sitzmann, Vincent and Sridhar, Srinath},
  booktitle={Computer graphics forum},
  volume={41},
  number={2},
  pages={641--676},
  year={2022},
  organization={Wiley Online Library}
}

@article{essakine2024we,
  title={Where do we stand with implicit neural representations? a technical and performance survey},
  author={Essakine, Amer and Cheng, Yanqi and Cheng, Chun-Wun and Zhang, Lipei and Deng, Zhongying and Zhu, Lei and Sch{\"o}nlieb, Carola-Bibiane and Aviles-Rivero, Angelica I},
  journal={arXiv preprint arXiv:2411.03688},
  year={2024}
}

@inproceedings{zhu2025large,
  title={Large images are Gaussians: High-quality large image representation with levels of 2D Gaussian splatting},
  author={Zhu, Lingting and Lin, Guying and Chen, Jinnan and Zhang, Xinjie and Jin, Zhenchao and Wang, Zhao and Yu, Lequan},
  booktitle={Proceedings of the AAAI Conference on Artificial Intelligence},
  volume={39},
  number={10},
  pages={10977--10985},
  year={2025}
}

@inproceedings{li2025gaussianimagepp,
  title={GaussianImage++: Boosted Image Representation and Compression with 2D Gaussian Splatting},
  author={Li, Tiantian and Zhang, Xinjie and Ge, Xingtong and Xu, Tongda and He, Dailan and Zhang, Jun and Wang, Yan},
  booktitle={Proceedings of the AAAI Conference on Artificial Intelligence},
  volume={40},
  number={8},
  pages={6442--6449},
  year={2026}
}

@article{liang2025structure,
  title={Structure-Guided Allocation of 2D Gaussians for Image Representation and Compression},
  author={Liang, Huanxiong and Chen, Yunuo and Pan, Yicheng and Wang, Sixian and Dai, Jincheng and Lu, Guo and Zhang, Wenjun},
  journal={arXiv preprint arXiv:2512.24018},
  year={2025}
}

@article{chen2026gifsplat,
  title={GIFSplat: Generative Prior-Guided Iterative Feed-Forward 3D Gaussian Splatting from Sparse Views},
  author={Chen, Tianyu and Xiang, Wei and Han, Kang and Lu, Yu and Wu, Di and Liu, Gaowen and Kompella, Ramana Rao},
  journal={arXiv preprint arXiv:2602.22571},
  year={2026}
}

@article{tancik2020fourier,
  title={Fourier features let networks learn high frequency functions in low dimensional domains},
  author={Tancik, Matthew and Srinivasan, Pratul and Mildenhall, Ben and Fridovich-Keil, Sara and Raghavan, Nithin and Singhal, Utkarsh and Ramamoorthi, Ravi and Barron, Jonathan and Ng, Ren},
  journal={Advances in neural information processing systems},
  volume={33},
  pages={7537--7547},
  year={2020}
}

@inproceedings{benbarka2022seeing,
  title={Seeing implicit neural representations as fourier series},
  author={Benbarka, Nuri and H{\"o}fer, Timon and Zell, Andreas and others},
  booktitle={Proceedings of the IEEE/CVF Winter Conference on Applications of Computer Vision},
  pages={2041--2050},
  year={2022}
}

@article{martel2021acorn,
  title={Acorn: Adaptive coordinate networks for neural scene representation},
  author={Martel, Julien NP and Lindell, David B and Lin, Connor Z and Chan, Eric R and Monteiro, Marco and Wetzstein, Gordon},
  journal={arXiv preprint arXiv:2105.02788},
  year={2021}
}

@inproceedings{lindell2022bacon,
  title={Bacon: Band-limited coordinate networks for multiscale scene representation},
  author={Lindell, David B and Van Veen, Dave and Park, Jeong Joon and Wetzstein, Gordon},
  booktitle={Proceedings of the IEEE/CVF conference on computer vision and pattern recognition},
  pages={16252--16262},
  year={2022}
}

@inproceedings{strumpler2022implicit,
  title={Implicit neural representations for image compression},
  author={Str{\"u}mpler, Yannick and Postels, Janis and Yang, Ren and Gool, Luc Van and Tombari, Federico},
  booktitle={European conference on computer vision},
  pages={74--91},
  year={2022},
  organization={Springer}
}

@article{lee2021meta,
  title={Meta-learning sparse implicit neural representations},
  author={Lee, Jaeho and Tack, Jihoon and Lee, Namhoon and Shin, Jinwoo},
  journal={Advances in Neural Information Processing Systems},
  volume={34},
  pages={11769--11780},
  year={2021}
}

@article{schwarz2022meta,
  title={Meta-learning sparse compression networks},
  author={Schwarz, Jonathan Richard and Teh, Yee Whye},
  journal={arXiv preprint arXiv:2205.08957},
  year={2022}
}

@article{dosovitskiy2020image,
  title={An image is worth 16x16 words: Transformers for image recognition at scale},
  author={Dosovitskiy, Alexey and Beyer, Lucas and Kolesnikov, Alexander and Weissenborn, Dirk and Zhai, Xiaohua and Unterthiner, Thomas and Dehghani, Mostafa and Minderer, Matthias and Heigold, Georg and Gelly, Sylvain and others},
  journal={arXiv preprint arXiv:2010.11929},
  year={2020}
}

@inproceedings{franchini2019stochastic,
  title={Stochastic Floyd-Steinberg dithering on GPU: image quality and processing time improved},
  author={Franchini, Giorgia and Cavicchioli, Roberto and Hu, Jia Cheng},
  booktitle={2019 Fifth International Conference on Image Information Processing (ICIIP)},
  pages={1--6},
  year={2019},
  organization={IEEE}
}

@article{simonyan2014very,
  title={Very deep convolutional networks for large-scale image recognition},
  author={Simonyan, Karen and Zisserman, Andrew},
  journal={arXiv preprint arXiv:1409.1556},
  year={2014}
}

@article{abdi2010principal,
  title={Principal component analysis},
  author={Abdi, Herv{\'e} and Williams, Lynne J},
  journal={Wiley interdisciplinary reviews: computational statistics},
  volume={2},
  number={4},
  pages={433--459},
  year={2010},
  publisher={Wiley Online Library}
}

@article{watson1981delaunay,
  title={Computing the n-dimensional Delaunay tessellation with application to Voronoi polytopes},
  author={Watson, David F.},
  journal={The Computer Journal},
  volume={24},
  number={2},
  pages={167--172},
  year={1981},
  doi={10.1093/comjnl/24.2.167}
}

@inproceedings{figurnov2017spatially,
  title={Spatially adaptive computation time for residual networks},
  author={Figurnov, Michael and Collins, Maxwell D and Zhu, Yukun and Zhang, Li and Huang, Jonathan and Vetrov, Dmitry and Salakhutdinov, Ruslan},
  booktitle={Proceedings of the IEEE conference on computer vision and pattern recognition},
  pages={1039--1048},
  year={2017}
}

@inproceedings{song2021variable,
  title={Variable-rate deep image compression through spatially-adaptive feature transform},
  author={Song, Myungseo and Choi, Jinyoung and Han, Bohyung},
  booktitle={Proceedings of the IEEE/CVF international conference on computer vision},
  pages={2380--2389},
  year={2021}
}

@inproceedings{minnen2017spatially,
  title={Spatially adaptive image compression using a tiled deep network},
  author={Minnen, David and Toderici, George and Covell, Michele and Chinen, Troy and Johnston, Nick and Shor, Joel and Hwang, Sung Jin and Vincent, Damien and Singh, Saurabh},
  booktitle={2017 IEEE International Conference on Image Processing (ICIP)},
  pages={2796--2800},
  year={2017},
  organization={IEEE}
}

@article{barron2022mipnerf360,
    title={Mip-NeRF 360: Unbounded Anti-Aliased Neural Radiance Fields},
    author={Jonathan T. Barron and Ben Mildenhall and 
            Dor Verbin and Pratul P. Srinivasan and Peter Hedman},
    journal={CVPR},
    year={2022}
}
\end{document}